\documentclass{article}

     \usepackage[preprint,nonatbib]{neurips_2019}

\usepackage[utf8]{inputenc} 
\usepackage[T1]{fontenc}    
\usepackage{hyperref}       
\usepackage{url}            
\usepackage{booktabs}       
\usepackage{amsfonts}       
\usepackage{nicefrac}       
\usepackage{microtype}      

 \usepackage{amsmath}
\usepackage{amsmath,amssymb,amsthm}
\usepackage{booktabs} 
\usepackage{amscd}
\usepackage{mathrsfs} 
\usepackage[shortlabels]{enumitem}
\usepackage{euscript}
\usepackage{graphicx}
\usepackage{amscd,bm}
\usepackage[utf8]{inputenc}
\usepackage[english]{babel}
\usepackage{multicol}
\usepackage{subfigure}
\usepackage{calrsfs}

\DeclareMathAlphabet{\pazocal}{OMS}{zplm}{m}{n}
\DeclareMathAlphabet\mathbfcal{OMS}{cmsy}{b}{n}

\usepackage{algorithm}
\usepackage{algorithmic}
\newtheorem{remark}{Remark}

\usepackage[square,sort,comma,numbers]{natbib}

\providecommand{\nor}[1]{\left\lVert {#1} \right\rVert}

\def\argmax{\operatornamewithlimits{arg\,max}}
\def\argmin{\operatornamewithlimits{arg\,min}}

\def\bit{\begin{itemize}}
\def\eit{\end{itemize}}

\usepackage{mathtools}

\def\ben{\begin{enumerate}}
\def\een{\end{enumerate}}

\usepackage{listings}
\usepackage{color}

\definecolor{dkgreen}{rgb}{0,0.6,0}
\definecolor{gray}{rgb}{0.5,0.5,0.5}
\definecolor{mauve}{rgb}{0.58,0,0.82}

\usepackage{enumitem}

\title{Wasserstein Style Transfer}

\author{%
  Youssef Mroueh\\
  IBM Research \\
  IBM T.J Watson Research Center\\
  \texttt{mroueh@us.ibm.com} \\
}

\begin{document}

\maketitle

\begin{abstract}
We propose Gaussian optimal transport for Image style transfer in an Encoder/Decoder framework  . Optimal transport for Gaussian measures has closed forms Monge mappings from source to target distributions. Moreover interpolates between a content and a style image can be seen as geodesics in the  Wasserstein Geometry. Using this insight, we show how to mix different target styles , using Wasserstein barycenter of Gaussian measures. Since Gaussians are closed under Wasserstein barycenter, this allows us a simple style transfer and style mixing and interpolation. Moreover we show how mixing different styles can be achieved using other geodesic metrics between gaussians such as the Fisher Rao metric, while the transport of the content to the new interpolate style is still performed with Gaussian OT maps. Our simple methodology allows to generate new stylized content interpolating between many artistic styles. The metric used in the interpolation results in different stylizations.

\end{abstract}

\section{Introduction}
Image style transfer consists in the task of modifying an image in a way that preserves its content and matches the artistic style of a target image or a collection of images. Defining a loss function that captures this content/style constraint is challenging.  A big progress in this field was made since the introduction of the neural style transfer in the seminal work of Gatys et al \citep{Gatys1,Gatys2}. Gatys et al showed that by matching statistics of the spatial distribution of images in the feature space of deep convolutional neural networks (spatial Grammian), one could define a style loss function. In Gatys et al method, the image is updated via an optimization process to minimize this ``network loss". One shortcoming of this approach is that is slow and that it requires an optimization per  content  and per style images. Many workarounds have been introduced to speedup this process via feedforward networks optimization that produce stylizations in a single forward pass \citep{Johnson_2016,UlyanovLVL16,LiW16,Wang_2017}. Nevertheless this approach was still limited to a single style image. \citep{Ulyanov_2017IN} introduced \emph{Instance Normalization} (IN) to improve quality and diversity of stylization. Multiple styles neural transfer was then introduced in \citep{dumoulin2017learned-iclr} thanks to  \emph{Conditional Instance Normalization} (CIN). CIN adapts the normalized statistics of the transposed convolutional layers in the feedforward network with learned scaling and biases for each style image for a fixed number of style images.  The concept of layer swap in   \citep{chen2016fast} resulted in one of the first arbitrary style transfer. \emph{Adaptive instance Normalization} was introduced in \citep{AdaIn} by making CIN scaling and biases learned functions from the style image, which enabled also arbitrary style transfer . The Whitening Coloring Transform (WCT)  \citep{WCT} which we discuss in details in Section \ref{sec:WCT} developed a simple framework for arbitrary style transfer using an Encoder/Decoder framework and operate a simple \emph{normalization} transform (WCT) in the encoder feature space to perform the style transfer. 

\noindent Our work is the closest to the WCT transform, where we start by noticing that instance normalization layers (IN,CIN, adaIN and WCT) are performing a transport map from the spatial distribution of a content image  to the one of a style image, and the implicit assumption in deriving those maps is the \emph{Gaussianity} of the spatial distribution of images in a deep CNN feature space. The Wasserstein geometry of Gaussian measures is very well studied in optimal transport \citep{takatsu2011} and Gaussian Optimal Transport (OT) maps have closed forms. We show in Section \ref{sec:approx} that those normalization transforms are approximations of the OT maps. Linear interpolations of different content or styles at the level of those normalization feature transforms have been successfully applied in \citep{AdaIn,dumoulin2017learned-iclr}  we show in Section \ref{sec:geodesic} that this can be interpreted and improved as Gaussian geodesics in the Wasserstein geometry . Furthermore using this insight, we show in Section \ref{sec:Interpolation} that we can define novel styles  using Wasserstein barycenyter of Gaussians \citep{Carlier}.  We also extend this to other Fr\'echet means in order to study the impact of the ground metric used on the covariances in the novel style obtained via this non linear interpolation. Experiments are presented in Section \ref{sec:exp}.

\section{Universal Style Transfer }\label{sec:WCT}
We review in this Section the approach of universal style transfer of WCT \citep{WCT}.

\noindent \textbf{Encoding Map.} Given a  content  image $I_c$ and a style image $I_s$ and a Feature extractor $F_j: \mathbb{R}^d \to \mathbb{R}^{m}, j=1\dots n,$ where $n$ is the spatial output of $F$, $m$ is its feature dimension . Define the following \emph{Encoding} map:
$\bm{E}: ~\mathbb{R}^d \to \mathcal{P}(\mathbb{R}^m): ~~  I\to \nu_{I}= \frac{1}{n}\sum_{j=1}^n \delta_{F_{j}(I)},$
\noindent
where  $\mathcal{P}(\mathbb{R}^m)$ is the space of empirical measures on $\mathbb{R}^m$. For example  $F$ is a VGG \citep{simonyan2014very} CNN that maps an image  to $\mathbb{R}^{C\times (H \times W)}$($C$ is the number of channels, $H$ the height and $W$ the width). In other words the CNN defines a distribution in the space of dimension $m=C$, and we are given $n=C \times H$ samples of this distribution. We note  $\nu=\bf{E}(I)$ this empirical distribution, i.e the spatial distribution of image $I_c$ in the feature space of a deep convolutional network $F$. 

\noindent \textbf{Decoding Map.} We assume that the encoding $\bf{E}$ is invertible , i.e exists:
$\bm{D}: ~ \mathcal{P}(\mathbb{R}^m) \to \mathbb{R}^d:   ~~  \nu \to  \bf{D}(\nu),$
such that $\bm{D}(\bm{E}(I))=I$. $\bm{(E, D)}$ is a VGG image Encoder/ Decoder for instance trained from the pixel domain to a spatial convolutional layer output in VGG and vice-versa. 

\noindent \textbf{Universal Style Transfer in Feature Space.} Universal style transfer approach  \citep{WCT} works in the following way: WCT (Whitening Coloring Transform) defines a transform $\bm{T}_{c\to s}$ (we will elaborate later on this transform) in the feature space $\mathbb{R}^m$: 
$\bm{T}_{c\to s}: ~ \mathbb{R}^{m} \to \mathbb{R}^m : ~~ x \to  \bm{T}_{c\to s}(x),$
the style transfer Transform $\bm{T}_{c\to s}$ operates in the feature space and defines naturally a push forward map on the spatial distribution of the features of content image $I_c$: $$ \bm{T}_{c\to s, \#}(\nu(I_c)):= \frac{1}{n}\sum_{j=1}^n \delta_{\bm{T}_{c\to s}(F_{j}(I_{c}))}. $$
$\bm{T}_{c\to s}$  is defined so that the style transfer happens in the feature space i.e $\bm{T}_{c\to s, \#}(\nu(I_c))= \nu(I_s)$. We obtain the stylized image $\tilde{I}_{c\to s}$ by decoding back to the image domain :
$$\tilde{I}_{c\to s} = \bm{D}(\textcolor{blue}{\bm{T}}_{c\to s,\#}(\bm{E}(I_c))).$$
From this formalism we see that the universal style transfer problem amounts to finding a transport map $\bm{T}_{c\to s}$  from the spatial distribution of a content image in a feature space $\nu(I_c)$ to the the spatial distribution of a target image in the same feature space $\nu(I_s)$. We show in the next section how to leverage optimal transport theory to define such maps. Moreover we show that the WCT transform and Adaptive instance normalization are approximations to the optimal transport maps.

\section{Wasserstein Universal Style Transfer} \label{sec:approx}
\noindent Given $ \nu_c= \bm{E}(I_c)$ and  $ \nu_s=\bm{E}(I_s)$, we formulate the style transfer problem as  finding an optimal Monge map: 
\begin{equation}
\inf_{T} \int \nor{x- T(x)}^2_2 d\nu_c(x) , \text{such that } T_{\#}(\nu_c)= \nu_s
\label{eq:Optimal}
\end{equation}
\noindent the optimal value of this problem is  $W^2_2(\nu_c,\nu_s)$, the  Wasserstein two distance between $\nu_c$ and $\nu_s$.
 Under some regularity conditions on the distributions, the optimal transport exists and is unique and $T_{c \to s }$ is the gradient of a convex potential \citep{dynamicTransport}


\noindent \textbf{Wasserstein Geometry of Gaussian Measures.} Computationally Problem \eqref{eq:Optimal} can be solved using for example entropic regularization of  the equivalent Kantorovich form of $W^2_2$ \citep{cuturi2013sinkhorn,peyre2017computational} or in an end to end approach using automatic differentiation of a Sinkhorn loss \citep{frogner2015learning,mmdSinkhorn} . We take here another route,
using the following known fact that Gaussian measures OT provides a lower bound on the Wasserstein distance \citep{Cuesta-Albertos1996} . For any two measures $\mu$ and $\nu$:
\vskip -0.1in
\begin{eqnarray*}
W^2_2(\mu, \nu) &\geq& W^2_2(\pazocal{N}(m_{\mu},\Sigma_{\mu}), \pazocal{N}(m_{\nu},\Sigma_{\nu}))\\
\end{eqnarray*}
\vskip -0.1in
\noindent where $m_{\mu},\Sigma_{\mu}$ are means and covariance of $\mu$, and $m_{\nu},\Sigma_{\nu}$ of $\nu$.
 The Wasserstein geometry of Gaussian measures is well studied and have many convenient computational properties \citep{takatsu2011}, we summarize them in the following:

\noindent  \emph{1) Closed Form $W^2_2$.} Given two Gaussians distributions $\nu= \pazocal{N}(m_{\mu},\Sigma_{\mu})$, and $\mu =\pazocal{N}(m_{\nu},\Sigma_{\nu})$ we have: 
$$ W^2_2(\pazocal{N}(m_{\mu},\Sigma_{\mu}), \pazocal{N}(m_{\nu},\Sigma_{\nu}))=\left|\left|m_{\nu}-m_{\nu}\right|\right|^2+ d^2_{\mathcal{B}}(\Sigma_{\mu},\Sigma_{\nu}), $$
where $$ d^2_{\mathcal{B}}(\Sigma_{\mu},\Sigma_{\nu})= \text{trace}\left(\Sigma_{\mu}+\Sigma_{\nu} - 2 \left( \Sigma^{\frac{1}{2}}_{\mu}\Sigma_{\nu}\Sigma^{\frac{1}{2}}_{\mu} \right)^{\frac{1}{2}}  \right)
$$ is the Bures metric between covariances. The Bures metric is a goedesic metric on the PSD cone. (In Section \ref{sec:FrechetMean} we discuss properties of this metric).

\noindent \emph{2) Closed Form Monge Map.} The optimal transport map between two  Gaussians with non degenerate covariances (full rank ) has a closed form:
$T_{\mu\to \nu}: x\to m_{\nu}+ A(x-m_{\mu}),$
where $A=\Sigma^{-\frac{1}{2}}_{\mu} \left(\Sigma^{\frac{1}{2}}_{\mu}\Sigma_{\nu}\Sigma^{\frac{1}{2}}_{\mu}\right)^{\frac{1}{2}}\Sigma^{-\frac{1}{2}}_{\mu}= A^{\top},$
i.e  $T_{\mu \to \nu, \#}(\mu)=\nu$ and $T_{\mu \to \nu}$ is optimal in the $W^2_2$ sense. If the Gaussian were degenerate we can replace the square root matrices inverses with pseudo-inverses \citep{PeyreTextureSynthesis}. 

\begin{figure*}[ht!]
	\centering	
	\includegraphics[width=\linewidth]{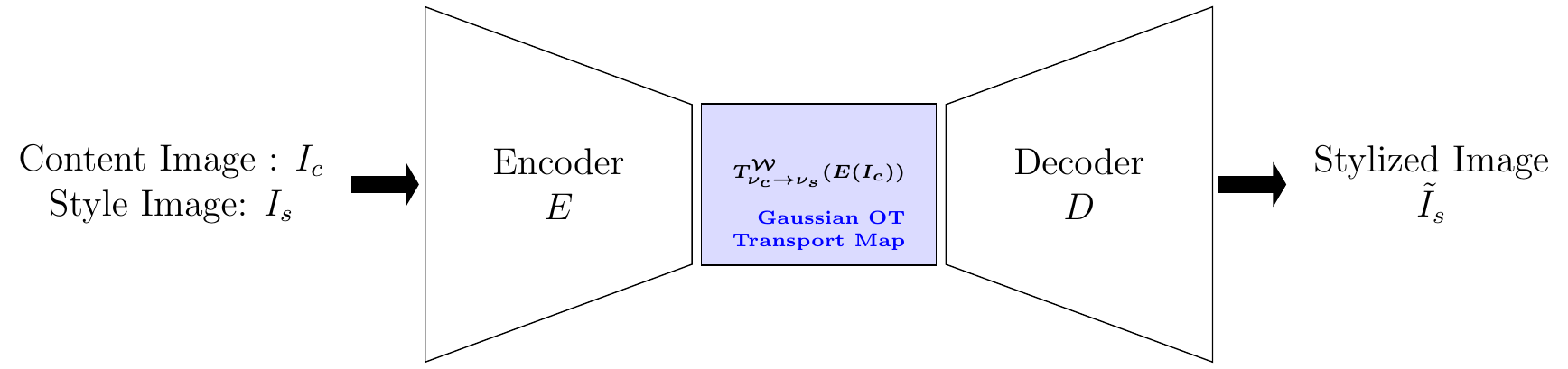} \\

	\caption{Wasserstein Style Transfer}
	 \label{fig:WST}
\end{figure*}

\noindent \textbf{Gaussian Wasserstein Style Transfer.} 
The spatial distribution of images in CNN feature space is not exactly Gaussian, but instead of having the solve Problem \eqref{eq:Optimal} we can use the Gaussian lower bound and obtain a closed form optimal map from the content distribution to the style distribution as follows: 
\begin{equation}
\boxed{\bm{T^{\pazocal{W}}}_{\nu_c\to \nu_s}(x)= \mu_s + A_{c\to s} (x - \mu_c ), }
\label{eq:WassGaussMap}
\end{equation}
where $\mu_{c}=\frac{1}{n}\sum_{j=1}^{n}F_j(I_c), \mu_{s}= \frac{1}{n} \sum_{j=1}^{n}F_j(I_s)$ , and $\Sigma_{c}= \frac{1}{n}\sum_{j=1}^{n} (F_{j}(I_c)-\mu_c)(F_{j}(I_c)-\mu_c)^{\top}$, and $\Sigma_{s}= \frac{1}{n}\sum_{j=1}^{n} (F_{j}(I_s)-\mu_s)(F_{j}(I_s)-\mu_s)^{\top}$ are means and covariances of $\nu_c$ and $\nu_s$ resp.
and $$A_{c\to s}=\Sigma^{-\frac{1}{2}}_{c} \left(\Sigma^{\frac{1}{2}}_{c}\Sigma_{s}\Sigma^{\frac{1}{2}}_{c}\right)^{\frac{1}{2}}\Sigma^{-\frac{1}{2}}_{c}.$$
Finally the Universal Wasserstein  Style Transfer can be written in the following compact way, that is summarized in Figure \ref{fig:WST}:
\begin{equation}
\boxed{\tilde{I}_{c\to s} = \bm{D}(\textcolor{blue}{\bm{T}^{\pazocal{W}}}_{\nu_c\to \nu_s,\#}(\bm{E}(I_c))).}
\end{equation}
\vskip -0.1 in

\noindent \textbf{Relation to WCT and to Adaptive Instance Normalization.}
We consider two particular cases: \\
\emph{1) Commuting  covariances and WCT \citep{WCT}.} Assuming that the covariances $\Sigma_c$ and $\Sigma_s$ commute meaning that $\Sigma_c\Sigma_s=\Sigma_s\Sigma_c$  ($\Sigma_s$ and $\Sigma_c$ have a common orthonormal basis ) it is easy to see that the optimal transport map reduces to : 
$$\bm{T^{\pazocal{W}}}_{\nu_c\to \nu_s}(x)= \mu_s +\Sigma_{s}^{\frac{1}{2}}  \Sigma^{-\frac{1}{2}}_c(x - \mu_c ) = \bm{T}^{\text{WCT}}_{c\to s}(x) $$
which is exactly the Whitening and Coloring Transform (WCT). Hence we see that WCT \citep{WCT} is only optimal when the covariances commute (a particular case is diagonal covariances). 

\noindent  \emph{2) Diagonal Covriances and AdaIN, Instance Normalization (IN) and Conditional Instance Normalization (CIN)\citep{AdaIn,Ulyanov_2017IN,dumoulin2017learned-iclr}}. Let $\sigma_{s}$ be the diagonal of $\Sigma_s$ and $\sigma_c$ be the diagonal of $\Sigma_c$. In case the covariances were diagonal it is easy to see that: 
$$\bm{T^{\pazocal{W}}}_{\nu_c\to \nu_s}(x)= \mu_s + \sqrt{\sigma_s} \odot \frac{(x-\mu_c)}{\sqrt{\sigma_c}} = \text{AdaIN}(x),$$
this is exactly the expression of adaptive instance normalization AdaIN. We conclude that AdaIN, IN and CIN implement a diagonal approximation of the optimal Gaussian transport map ($(\mu_s,\sigma_s)$, are learned constant scaling and biases in IN and CIN , and are adaptive in adaIN).

\section{ Wasserstein Style/Content Interpolation with McCann Interpolates} \label{sec:geodesic} 
One shortcoming of the formulation in problem \eqref{eq:Optimal} is that it does not allow to balance the content/style preservation as it is the case in end to end style transfer. Let $t \in [0,1]$ we formulate the style transfer problem with content preservation as follows:
\begin{equation}
\min_{\nu} (1-t) W^2_2(\nu, \nu_c)+  t W^2_2(\nu,\nu_s),
\label{eq:styletransfer}
\end{equation}
The first term in Equation \eqref{eq:styletransfer} measure the usual "content loss" in style transfer and the second term measures the   "style loss". $t$ balances the interpolation between the style and the content. 
In optimal transport theory, Problem \eqref{eq:styletransfer} is known as the McCann interpolate \citep{MCCANN1997} between $\nu_c$ and $\nu_s$ and the solution of
\eqref{eq:styletransfer}  is a Wasserstein geodesic from $\nu_c$ to $\nu_s$ and is given by:
$$\nu_{t}=[(1-t) \text{Id}+ tT_{\nu_c\to \nu_s}]_{\#}(\nu_c)$$  
The spatial distribution of images in CNN is not exactly Gaussian, but instead of having the solve Problem \eqref{eq:styletransfer} we can again use the following  Gaussian lower bound:
\begin{equation}
\min_{\nu \sim \pazocal{N}(\mu, \Sigma)}(1- t) W^2_2( \pazocal{N}(\mu,\Sigma), \pazocal{N}(\mu_c,\Sigma_c))+ t W^2_2( \pazocal{N}(\mu,\Sigma), \pazocal{N}(\mu_s,\Sigma_s)).
\end{equation}
Fortunately this problem has also  a closed form \citep{MCCANN1997}: 
$$ \nu_{t}= \pazocal{N}(\mu_t, \Sigma_t)=[(1-t) \text{Id}+ t\bm{T}^{\pazocal{W}}_{\nu_c\to \nu_s}] ]_{\#}(\nu_c) ,$$
where $\bm{T}^{\pazocal{W}}_{\nu_c\to \nu_s}$ is given in Equation \eqref{eq:WassGaussMap}.
$\{\nu_{t}\}_{t\in [0,1]}$ is a geodesic between $\nu_c$ and $\nu_s$.
Finally the Wasserstein  Style/Content  Interpolation can be written in the following compact way: 
\begin{equation}
\boxed{\nu_{t}= (1-t) \bm{E}(I_c)+ t  \textcolor{blue}{\bm{T}^{\pazocal{W}}}_{\nu_c\to \nu_s,\#}(\bm{E}(I_c)),   ~~
\tilde{I}^t_{c\to s} = \bm{D}(\nu_t ).}
\end{equation}
In practice both WCT and AdaIN propose similar interpolations in feature space, we give here a formal justification for this approach. This formalism allows us to generalize to multiple styles interpolation using the  Gaussian Wasserstein geometry of the spatial distribution of CNN images features.  

\section{ Wasserstein Style Interpolation }\label{sec:Interpolation}
Given $\{(I^{j}_{s},\lambda_j)\}_{j=1\dots S}$, $S$ target styles images, and a content image $(I_c,\lambda_{S+1})$,where $\lambda_j$ are interpolation factors such that $\sum_{j=1}^{S+1}\lambda_j=1$. A naive approach to content/$S$ styles interpolation can be given by:
\vskip -0.2in 
$$\nu_{\lambda}=\sum_{j=1}^S \lambda_j \bm{T}^{\pazocal{W}}_{c\to s_j,\#}(\bm{E}(I_c))+ \lambda_{S+1} \bm{E}(I_c), ~~ I^{\lambda}_s= \bm{D}(\nu_{\lambda}),$$
\vskip -0.2in 
\noindent this approach was proposed in both WCT and AdaIn by replacing $\bm{T}^{\pazocal{W}}$ by $T^{\text{WCT}}$ and  AdaIN respectively. We show here how to define a non linear interpolation that exploits the Wasserstein geometry of Gaussian measures.  
\begin{figure}[ht!]
\centering
\includegraphics[scale=0.40]{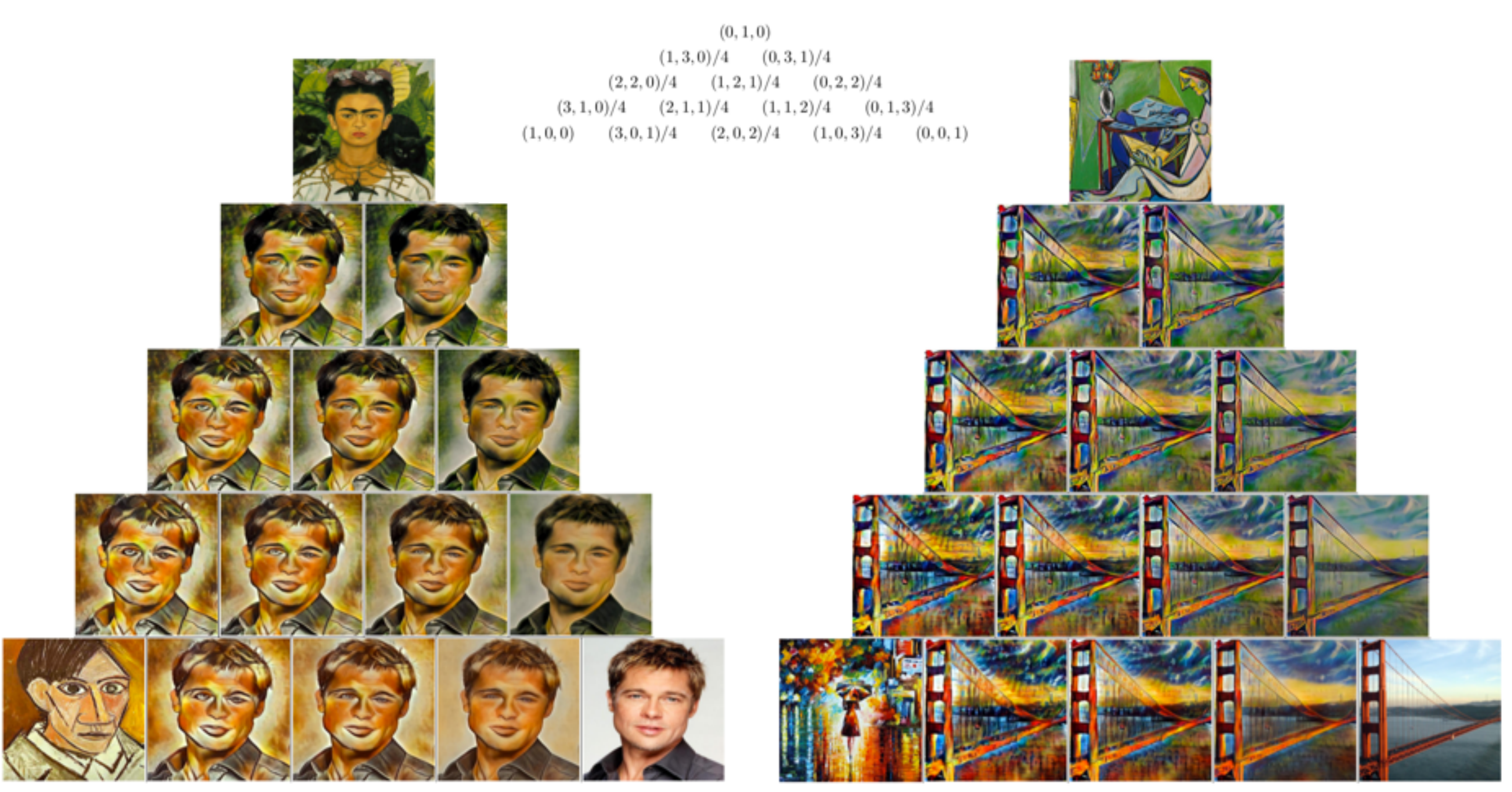}
\caption{Wasserstein Barycenter Interpolation between a content image and two target styles images. The weights $\{\lambda_j\}$ used are given above the two examples.}
\label{fig:InterpContentStyle}
\end{figure}

\subsection{Interpolation with Wasserstein Barycenters}
Similarly to the content /style interpolation, we formulate the content / $S$ styles interpolation problem as a Wasserstein Barycenter problem \citep{Carlier} as follows. Let $\nu^j_s= \bm{E}(I^j_{s})$, and $\nu_c= \bm{E}(I_c)$, we propose to solve the following Wasserstein Barycenter problem:
\vskip -0.2in
$$\nu^s_{\lambda}=\argmin_{\nu} \sum_{j=1}^{S} \lambda_j W^2_2(\nu,\nu^j_s) +\lambda_{S+1}W^2_2(\nu,\nu_c) $$ 
and then find the optimal map from $\nu_c$ to  the barycenter measure $\nu^s_\lambda$ $ T_{\nu_c\to \nu^s_{\lambda}}.$ The final stylized image is obtained as follows:
$\tilde{I}^{\lambda}_s= \bm{D} (\bm{T}_{\nu_c\to \nu^s_{\lambda}}(\bm{E}(I_c)) ) .$

\noindent Again we resort to Gaussian optimal transport lower bound of the above problem: 
\vskip -0.3 in 
\begin{equation}
\nu^s_{\lambda}=\argmin_{\nu \sim \pazocal{N}(\mu, \Sigma)}\sum_{j=1}^S \lambda_j W^2_2( \pazocal{N}(\mu,\Sigma), \pazocal{N}(\mu^j_{s},\Sigma^j_{s}))+ \lambda_{S+1} W^2_2( \pazocal{N}(\mu,\Sigma), \pazocal{N}(\mu_c,\Sigma_c)),
\end{equation}
As shown by Agueh and Carlier \citep{Carlier} the Wasserstein Barycenter of Gaussians is itself a Gaussian $\nu^{s}_{\lambda}= \pazocal{N}(\bar{\mu}_{\lambda},\bar{\Sigma}_{\lambda}),$
where $\bar{\mu}_{\lambda}= \sum_{j=1}^S \mu^s_j+ \lambda_{S+1} \mu_{c},$
and $\bar{\Sigma}_{\lambda}$ is a Bures Mean. Noting $\Sigma^{S+1}_s=\Sigma_c$ we have:
\vskip -0.3 in
$$ \bar{\Sigma}_{\lambda}=\argmin_{\Sigma} \sum_{j=1}^{S+1} \lambda_j d^2_{\mathcal{B}}(\Sigma,\Sigma^j_s)$$
 Agueh and Carlier showed that $\bar{\Sigma}_{\lambda}$ is the unique positive definite matrix solution of the following fixed point problem:
 $\bar{\Sigma}_{\lambda} = \sum_{j=1}^{S+1} \lambda_j \left(\bar{\Sigma}^{\frac{1}{2}} \Sigma^j_s \bar{\Sigma}^{\frac{1}{2}} \right)^{\frac{1}{2}}.$
 In order to solve this problem we use an alternative fixed point strategy proposed in \citep{algWass}, since it converges faster in practice:
 \begin{equation}
 \bar{\Sigma}_{\ell}= \sum_{j=1}^{S+1} \lambda_j  \bar{\Sigma}_{\ell-1}^{-\frac{1}{2}}\left(\bar{\Sigma}^{\frac{1}{2}}_{\ell-1} \Sigma^j_s \bar{\Sigma}^{\frac{1}{2}}_{\ell-1} \right)^{\frac{1}{2}} \bar{\Sigma}_{\ell-1}^{-\frac{1}{2}}, \ell = 0\dots L-1, 
 \label{eq:BuresBary}
 \end{equation}
 \vskip -0.2 in
 \noindent and we initialized as in \citep{PeyreTextureSynthesis}:
  $\bar{\Sigma}_0 = \Sigma^{j_0}_s, j_0=\argmax_{j=1\dots S+1}\lambda_j ,$
we found that $L=50$ was enough for convergence, i.e we set $\bar{\Sigma}_{\lambda}=\bar{\Sigma}_{L}$. Matrix square root and inverses were computed using SVD which gives an overall complexity of $O(Lm^3)$ and we used truncated SVD to stabilize  the inverses. 
Finally since the Barycenter is a Gaussian , the optimal transport map from  the  Gaussian spatial content distribution $\pazocal{N}(\mu_c,\Sigma_c)$ to the barycenter (mix of styles and content) $\pazocal{N}(\bar{\mu}_{\lambda},\bar{\Sigma}_{\lambda})$ is given in closed form as in Equation \eqref{eq:WassGaussMap}:
\begin{equation}
\boxed{\bm{T}^{\mathcal{W}}_{\nu_c \to \nu^s_{\lambda}}(x)= \bar{\mu}_{\lambda}+ \Sigma^{-\frac{1}{2}}_{c} \left(\Sigma^{\frac{1}{2}}_{c}\bar{\Sigma}_{\lambda}\Sigma^{\frac{1}{2}}_{c}\right)^{\frac{1}{2}}\Sigma^{-\frac{1}{2}}_{c}.(x-\mu_c)}.
\label{eq:content2bary}
\end{equation}
Finally to obtain the stylized image as a result of targeting the mixed/style $\nu^s_{\lambda}$ we decode back:
$$\tilde{I}^{\lambda}_{cs}=\bm{D}(\bm{T}^{\mathcal{W}}_{\nu_c \to \nu^s_{\lambda}}(\nu_c)).$$

\noindent Figure \ref{fig:InterpContentStyle} gives an example of our approach for mixing content images with style images. We see that the Wasserstein barycenter captures not only the color distribution but also the details of the artistic style (for instance Frida Kahlo's unibrow is well captured smoothly in the transition between Picasso self portrait and Frida Kahlo).

\subsection{Style Interpolation with Fr\'echet Means}\label{sec:FrechetMean}
In the previous section we defined interpolations between the content and the styles images. In this section we define  a "novel style" via an interpolation of style images only, we then map the content  to the novel style using Gaussian optimal transport. 

\noindent \textbf{From Wasserstein Barycenter to Fr\'echet Means on the PSD manifold.} As discussed earlier the Wasserstein Barycenter of the Gaussian approximations of the spatial distribution of style images in CNN feature spaces can be written as:
\vskip -0.2in
\begin{equation}
\min_{\mu, \Sigma} \sum_{j=1}^{S} \lambda_j \left(d^2_{\mu}(\mu,\mu^j_s)+ d^2_{\text{cov}}(\Sigma,\Sigma^{j}_s)\right), 
\label{eq:avStyle}
\end{equation}
\vskip -0.2in
\noindent for $d^2_{\mu}(\mu,\mu')=\nor{\mu-\mu'}^2_2$ the euclidean metric , $d^2_{cov}(\Sigma,\Sigma')= d^2_{\mathcal{B}}(\Sigma,\Sigma'),$
the Bures metric. The Bures Metric is a geodesic metric on the positive definite cone and and has another representation as a procrustes registration metric \citep{masarotto2018procrustes}:
\vskip -0.24in
 $$d^2_{\mathcal{B}}(\Sigma, \Sigma')=\min_{U\in \mathbb{R}^{m\times m} , UU^{\top}=I}\nor{\sqrt{\Sigma}- \sqrt{\Sigma'}U}^2_{F}.$$
 \vskip -0.1in
\noindent From this we see the advantage of Wasserstein barycenter on for example using $d^2_{\text{cov}}(\Sigma,\Sigma')=\nor{\Sigma- \Sigma'}^2_{F} $ the Frobenius norm. Bures Metric aligns the the square root of covariances using a rotation.
From this we see that by defining a new metric on covariances we can get different form of interpolates, we fix $d^2_{\mu}(\mu,\mu')=\nor{\mu-\mu'}^2_2$, and hence on $\mu$ we use always the arithmetic mean $\mu_{arth}=\sum_{j=1}^{S}\lambda_j \mu^j_s$. We give here different metrics $d^2_{cov}$ that defines different  Fr\'echet means on the PSD manifold (see \citep{Bhatia2013} and references there in )

\noindent \emph{\textbf{1) Arithmetic Mean:}} Solving Eq. \eqref{eq:avStyle} for $d^2_{\text{cov}}(\Sigma,\Sigma')= \nor{\Sigma-\Sigma'}^2_{F}$, we define the target style $\nu^{\lambda,\text{arth}}_{s}= \pazocal{N}(\mu_{arth},\Sigma_{\text{arth}})$, where $\Sigma_{\text{arth}}=\sum_{j=1}^{S}\lambda_j \Sigma^s_j$.\\
\noindent \emph{\textbf{2) Harmonic Mean:}}  Solving Eq. \eqref{eq:avStyle} for $d^2_{\text{cov}}(\Sigma,\Sigma')= \nor{\Sigma^{-1}-\Sigma'^{-1}}^2_{F}$, we define the target style $\nu^{\lambda,\text{Harm}}_{s}= \pazocal{N}\left(\mu_{\text{arth}},\Sigma_{\text{Harm}}\right)$, where $ \Sigma_{\text{Harm}}=\left(\sum_{j=1}^{S}\lambda_j (\Sigma^{s}_j)^{-1}\right)^{-1}$.\\ 
\noindent \emph{\textbf{3) Fisher Rao Mean (Karcher or Geometric Mean)}}. For $d^2_{cov}(\Sigma,\Sigma')= \nor{\log(\Sigma^{-\frac{1}{2}}\Sigma'\Sigma^{-\frac{1}{2}})}^2_{F}= 2\mathcal{F}^2(\pazocal{N}(0,\Sigma),\pazocal{N}(0,\Sigma')),$ that is the Riemannian natural metric or the Fisher Rao metric between Centered Gaussians. $\log$ here refers to matrix logarithm. The Fisher Rao metric  is a geodesic distance and its metric tensor is the Fisher information matrix . 
 Solving Eq. \eqref{eq:avStyle} with the Fisher Rao metric we obtain the so called Karcher Mean between PSD matrices $\Sigma_{\text{FisherRao}}$, and we define the target style  $\nu^{\lambda,\text{FisherRao}}_{s}= \pazocal{N}\left(\mu_{\text{arth}},\Sigma_{\text{FisherRao}}\right)$. \\
In order to find the Karcher mean we use manifold optimization techniques of  \citep{Zhang2016FirstorderMF} as follows. The gradient manifold update is :
\begin{equation}
\Sigma_{\ell}=\Sigma^{\frac{1}{2}}_{\ell-1}\exp \left(-\eta \sum_{j=1}^S \log \left(\Sigma^{\frac{1}{2}}_{\ell-1}(\Sigma^{s}_{j})^{-1}\Sigma^{\frac{1}{2}}_{\ell-1}\right)\right)\Sigma^{\frac{1}{2}}_{\ell-1},
\label{eq:FisherRaoBary}
\end{equation}
we initialize $\Sigma_0$ as in the Wasserstein case and iterate for $L=50$ iterations with $\eta$ the learning rate set to $0.01$.

\begin{remark}
While we defined here the barycenter style of each metric as a Gaussian, Wasserstein Barycenter is the only one  that guarantees a Gaussian barycenter \citep{Carlier}.
\end{remark}

\noindent \textbf{Mapping  a content image to a target novel style.} Given now the new style $\nu^{\lambda,\text{mean}}_s$, where mean is in $\{\text{arth}, \text{harm}, \text{Fisher Rao}, \text{Wasserstein}\}$, we stylize a content image $I_c$ using Gaussian Optimal transport as described in the paper:
$$\boxed{\tilde{I}_{c\to s} = \bm{D}(\textcolor{blue}{\bm{T}^{\pazocal{W}}}_{\nu_c\to \nu^{\lambda, \text{mean}}_s,\#}(\bm{E}(I_c))).}
$$
Our approach is summarized in Algorithms \ref{alg:styleBary} and \ref{alg:Bary} given in Appendix. 
\section{Related works }
\textbf{OT for style Transfer and Image coloring.} Color transfer between images using regularized optimal transport on the  color distribution of images (RGB for example) was studied and applied in  \citep{Ferradans_2013}.  The color distribution is not gaussian and hence the OT problem has to be solved using regularization. Optimal transport for style transfer using the spatial  distribution in the feature space of a deep CNN was also explored in  \citep{OTstyle,kolkin2019style}. \cite{OTstyle} uses $W^2_2$ for Gaussians  as content and style loss and optimizes it in an end to end fashion similar to \citep{Gatys1,Gatys2}. \citep{kolkin2019style} uses an approximation of the Wasserstein distance as a loss that is also optimized in an end to end fashion.  Both approaches don't allow universal style transfer and an optimization is needed for every style/content image pairs.

\noindent \textbf{Wasserstein Barycenter for Texture Mixing. } Similar to our approach for Wasserstein mixing in an encoder/decoder framework, \citep{PeyreTextureWass} uses the wavelet transform to encode textures, applies Wasserstein barycenter on wavelets coefficients, and then decodes back using the inverse wavelet transform to synthesize a novel mixed texture. The wasserstein barycenter problem there has to be solved exactly and the Gaussian approximation can not be used since the wavelet coffecient distribution is not Gaussian. A special model for Gaussian texture mixing was developed in \citep{PeyreTextureSynthesis}. The advantage of using features of a CNN is that the Gaussian lower bound of the Wasserstein distance seems to be tight.  

\noindent \textbf{Other approaches to style Transfer.} While our focus in this paper was on OT metrics for style transfer other approaches exist (see  \citep{reviewStyle} for a review) and have used different type of losses such as MRF loss \citep{Li_2016} , MMD loss \citep{MMDLi_2017},  GAN loss \citep{LiW16} and cycle GAN loss \citep{CycleGAN2017}.  
%

\section{Experiments}\label{sec:exp}

In order to test our approach of geometric mixing of styles we use the WCT framework \citep{WCT}, where we use a pyramid of $5$ encoders $(E_r,D_r), r=1\dots 5$ at different  spatial resolutions, where $(E_5,D_5)$ corresponds to the coarser resolution, and $(E_1,D_1)$  the finer resolution. Following WCT we use a coarse to fine approach to style transfer as follows. Given interpolation weights $\{\lambda_j, j=1\dots S\}$, we start with $r=5$ and with $\nu_c= E_{5}(I_c)$:
\begin{figure}[bt!]
\centering
\includegraphics[scale=0.40]{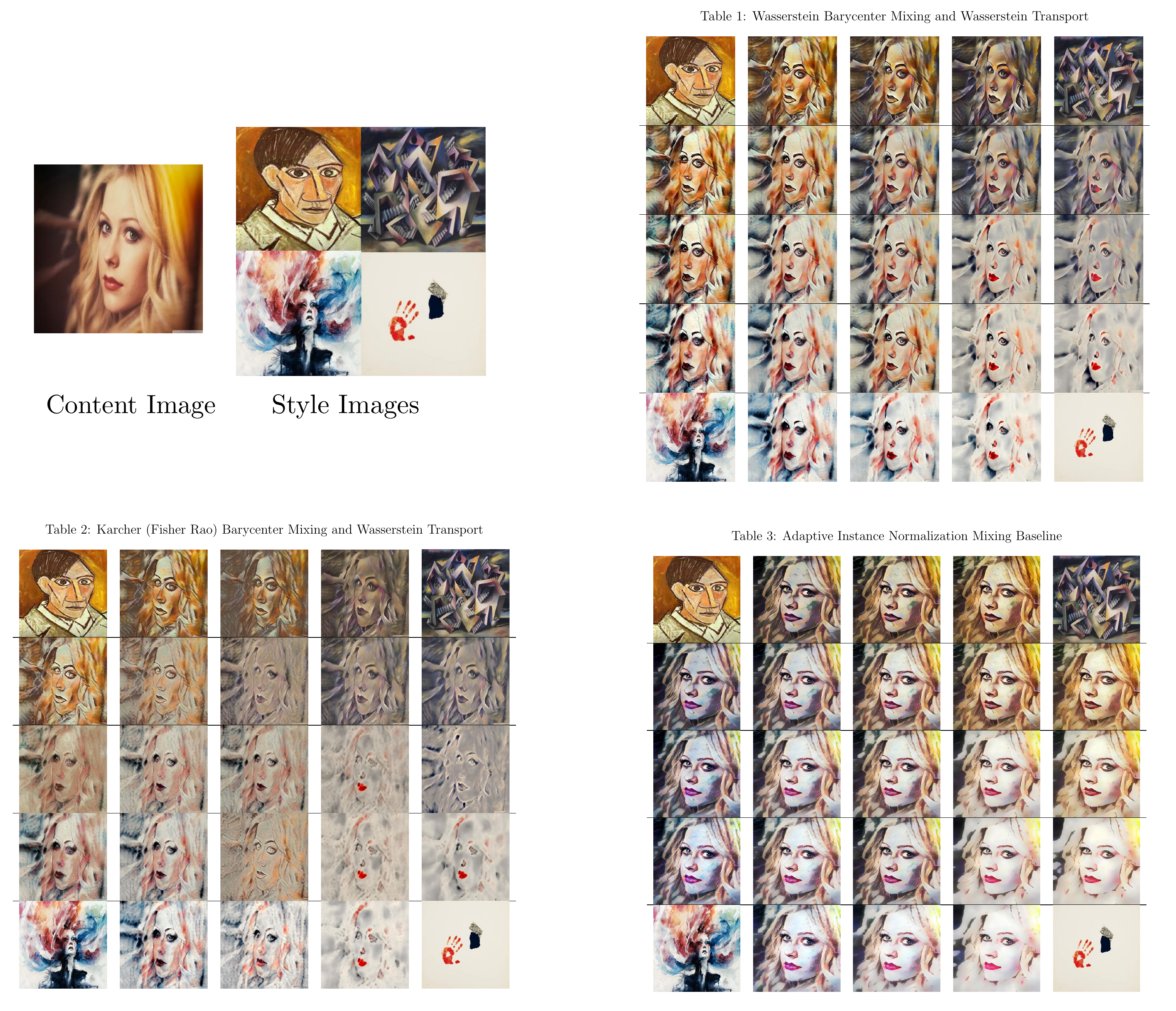}
\caption{(
Table 1): Wasserstein Barycenter Interpolation between a content image given above and four target styles images given at the corner of the square.  Each image in the square is for an interpolation weight $(\lambda_1,\dots \lambda_4)$, that are defined on a grid on the square. (Table 2): Fisher Rao Interpolation between the same content image given above and the same four target styles images given at the corner of the square. In both cases Gaussian Wasserstein transport plans are used to obtain the transformed image to the novel mixed style in the feature space, and the final image is obtained using the decoder. (Table 3): the AdaIn baseline that we showed that it does a diagonal approximation fails at capturing the subtle details of the style of the target images.  Both Wasserstein and Fisher Rao approaches are successful, we notice that while Wasserstein barycenter is color dominant in defining the new style, the Fisher Rao barycenter capture more the strokes and captures better color variations in the novel artistic styles.  We note that the Wasserstein is smoother as we change the interpolation weights then the Fisher Rao. (Figure is better seen in color and zooming in; See Appendix for a full resolution). }
\label{fig:MixingGoedesic}
\vskip -0.2in
\end{figure}
\begin{enumerate}
\item We encode all style images at resolution $r$, $E_{r}(I^j_s),j=1\dots s$. We define the mixed style $\nu^{\lambda,r}_s$ at resolution $r$ using one of the mixing strategies (Frechet Mean) in Section \ref{sec:geodesic}, using Algorithms \ref{alg:styleBary} and \ref{alg:Bary}.
\item We find the Wasserstein Transport map at resolution $r$ between  the content $\nu_c$ and the novel style $\nu^{\lambda,r}_s$  and compute the transformed features:  $\nu^{r}_{cs}=\bm{T}^{\pazocal{W}}_{\nu_c\to \nu^{\lambda,r}_s,\#}(\nu_c)$.
\item We decode the novel image at resolution $r$ :  $I^{r}_{c}= D_r(\nu^{r}_{cs})$.
\item We set $\nu_c = E_{r-1}(I^{r}_{c})$, then set $r$  to $r-1$ and go to step $1$ until reaching $r=1$.
\end{enumerate}
The  stylized output of this procedure is  $I^{1}_{c}$. We also experimented in the Appendix with applying the same approach but in fine to coarse way starting from the higher resolution $r=1$ to the lower resolution encoder $r=5$. We show in Figure \ref{fig:MixingGoedesic} the output of our mixing strategy using two of the geodesic metrics namely Wasserstein and Fisher Rao barycenters. We give as baseline the AdaIn output for this (this same example was given in \citep{AdaIn} we reproduce it using their available code). We show that using geodesic metrics to define the mixed style successfully capture the subtle details of different styles.  More examples and comparison to literature and other types of mixing can be found in the Appendix.

\section{Discussion and Conclusion }
We conclude this paper by the following three observations on the spatial distribution of features in a deep convolutional neural network:
\begin{enumerate}
\item The success of Gaussian optimal transport between spatial distributions of deep CNN features that we demonstrated in this paper suggests that the network learned to "gaussianize" the space. Gaussinization \citep{gaussinization} is a principle in unsupervised learning. It will be interesting to further study this Gaussianity hypothesis and to see if Gaussinization can be used as a regularizer for learning deep CNN or as an objective in self-supervised learning.   
\item We showed that many of the spatial  normalization layers  used in deep learning such as Instance normalization \citep{Ulyanov_2017IN} and related variants can be understood as approximations of Gaussian optimal transport. When used in an architecture between layers, the normalization layer acts like a transport map between the spatial distribution of consecutive layers. We hope this angle will help developing new normalization layers and a better understanding of the existing ones. 
\item Geodesic metrics such as Wasserstein and the Fisher Rao metric allow better non linear interpolation in feature space.
\end{enumerate}

\newpage 
\bibliographystyle{unsrt}
\bibliography{simplex,refs}
\newpage
\appendix
\begin{center}
\textbf{Supplementary Material for Wasserstein Style Transfer }
\end{center}
\section{Algorithms }
\begin{center}
\vskip -0.2in
\begin{minipage}[t]{0.49\textwidth}
\null 
\begin{algorithm}[H]
\caption{\textsc{Frechet Mean Style Interpolation and Content Stylization}($d_{cov}$)}
\begin{algorithmic}
 \STATE {\bfseries Inputs:} $\{I^j_s\}_{j=1\dots S}$ style images, content Image $I_c$ , interpolations weights $\{\lambda_j\}_{j=1\dots S+1}$, Encoder/Decoder $(\bm{E},\bm{D})$,. \\
 \STATE {\bfseries Encode:}  $ \nu_c = \bm{E}(I_c), \nu^j_{s}= \bm{E}(I^j_s), j=1\dots S$ \\
 \STATE {\bfseries Statistics:}  $(\mu_c,\Sigma_c), (\mu^j_s,\Sigma^j_s), j=1\dots S$ \\
 \STATE {\bfseries Content/Style or Style only:} if  content/style $\mu^{s}_{S+1}=\mu_c, \Sigma^{s}_{S+1}=\Sigma_c, ,S\gets S+1$ , else pass.
 \STATE   {\bfseries Target Bary Mean:}  $\bar{\mu}_{\lambda}= \sum_{j=1}^{S}\lambda_j \mu_j$.\\
\STATE   {\bfseries Target Bary Covariance:}  $\bar{\Sigma}_{\lambda} =$\textsc{Frechet mean}$(\{\lambda_j,\Sigma^j_s\}, d_{cov})$\\
\STATE {\bfseries Novel Style:} $\nu^{\lambda}_{s}=\pazocal{N}(\bar{\mu}_{\lambda},\bar{\Sigma}_{\lambda})$\\
\STATE  {\bfseries   Gaussian OT Content to Target:} Compute $\nu_{cs}=\bm{T}^{\mathcal{W}}_{\nu_c \to \nu^s_{\lambda},\#}(\nu_c)$ given in Eq. \eqref{eq:content2bary}

\STATE {\bfseries Decode:} $\bm{D}(\nu_{cs})$
 \end{algorithmic}
 \label{alg:styleBary}
\end{algorithm}
\end{minipage}%
\hfill
\begin{minipage}[t]{0.49\textwidth}
\null
\begin{algorithm}[H]
\caption{\textsc{FRECHET MEAN}($\{\lambda_j,\Sigma^j_s\},d_{\text{cov}})$}
\begin{algorithmic}
 \STATE {\bfseries Initialize:} $\bar{\Sigma}_0 = \Sigma^{j_0}_s, j_0=\argmax_{j=1\dots S}\lambda_j $\\
 \STATE{\bfseries  if $d_{cov}= d_{\text{Bures}}$} find $\bar{\Sigma}_{\lambda}$ solve using iterations in Eq \eqref{eq:BuresBary}\\
  \STATE{\bfseries  if $d_{cov}= d_{\text{Fisher Rao}}$} find $\bar{\Sigma}_{\lambda}$ solve using iterations in Eq \eqref{eq:FisherRaoBary}
  \STATE{\bfseries  if $d_{cov}= d_{\text{Frobeinus}}$} $\bar{\Sigma}_{\lambda}= \sum_{j}\lambda_j \Sigma^s_j$
 \STATE {\bfseries  if $d_{cov}= d_{\text{Harmonic}}$}  $\bar{\Sigma}_{\lambda}= (\sum_{j}\lambda_j (\Sigma^s_j)^{-1})^{-1}$
 \end{algorithmic}
 \label{alg:Bary}
\end{algorithm}
\end{minipage}
\end{center}

\section{Examples of Interpolating Content and Styles with Wasserstein Barycenter and Optimal Transport}
\begin{figure}[ht!]
 \includegraphics{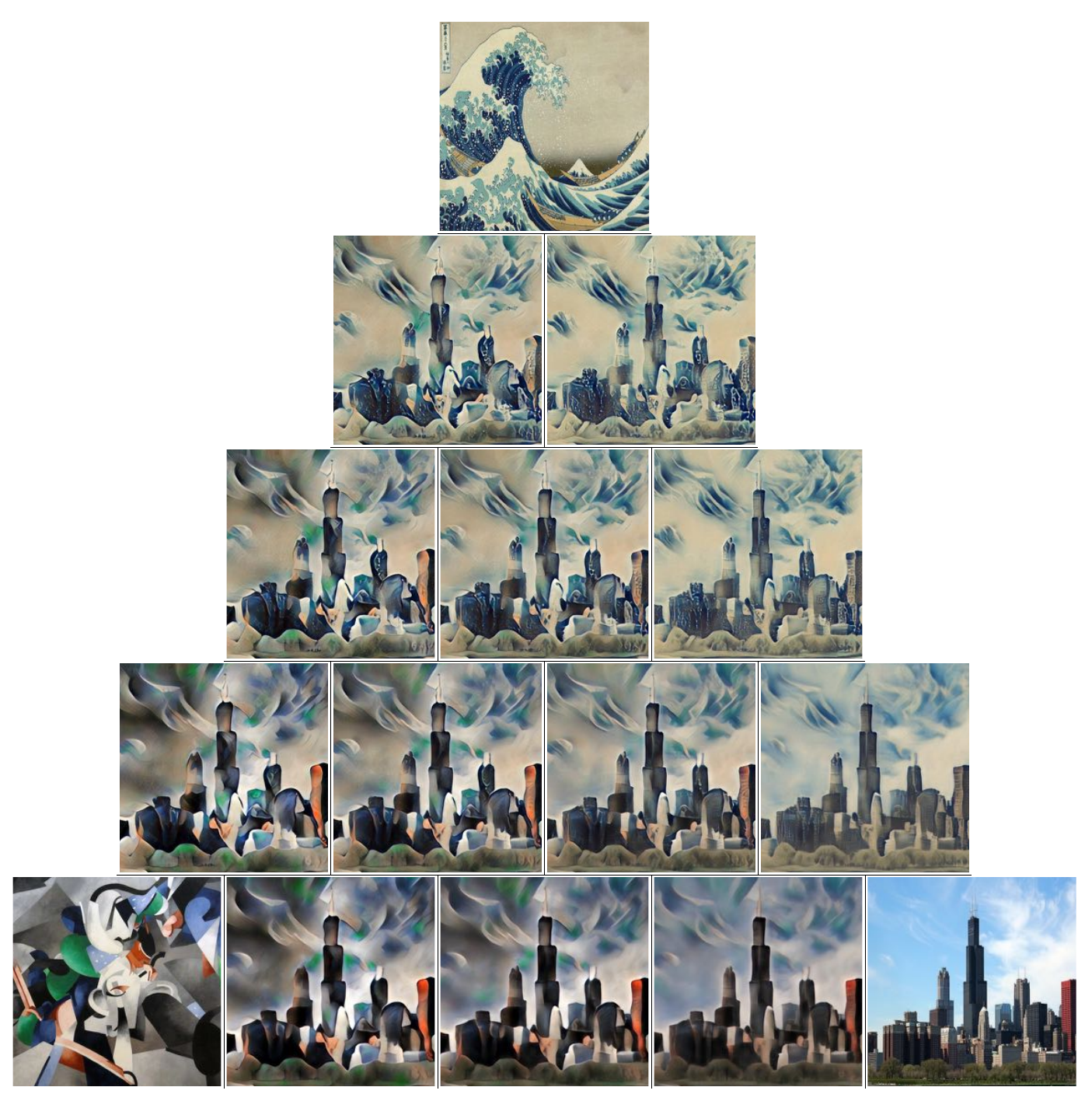}
\caption{Wasserstein barycenters for Style Mixing and Transfer. The content image on the right corner of the triangle is mixed with the two styles images. Each image in the triangle correspond to a set of  interpolation weights defined by proximity to the content or style images in the triangle.  }
\label{fig_wasserstein_triangle}
\end{figure}

\begin{figure}[ht!]
 \includegraphics{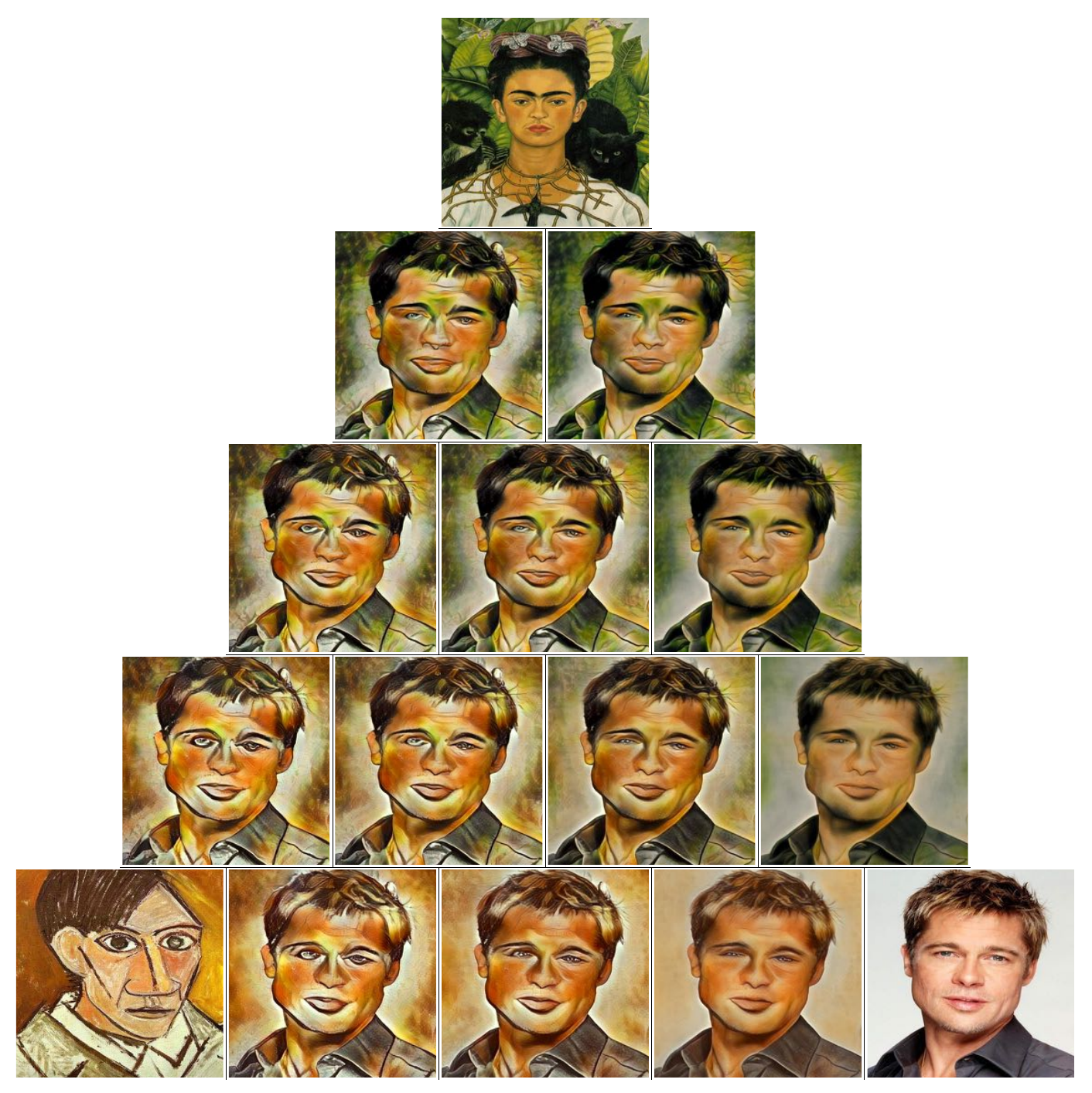}
\caption{Wasserstein barycenters for Style Mixing and Transfer. The content image on the right corner of the triangle is mixed with the two styles images. Each image in the triangle correspond to a set of  interpolation weights defined by proximity to the content or style images in the triangle.}
\label{fig_wasserstein_triangle2}
\end{figure}

\begin{figure}[ht!]
  \includegraphics{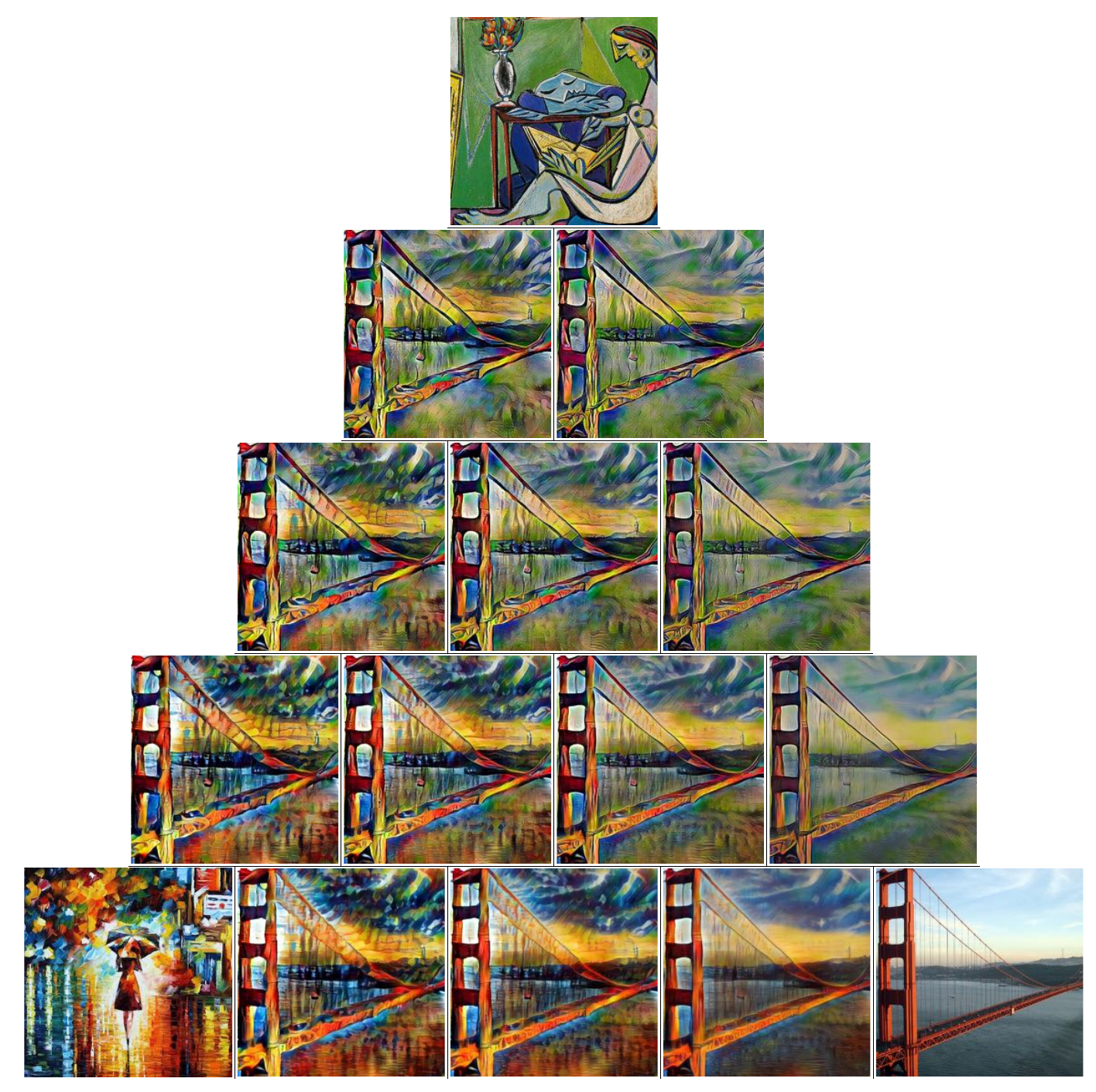}
 \caption{Wasserstein barycenters for Style Mixing and Transfer. The content image on the right corner of the triangle is mixed with the two styles images. Each image in the triangle correspond to a set of  interpolation weights defined by proximity to the content or style images in the triangle.}
\label{fig_wasserstein_triangle3}
\end{figure}

\noindent In Figures \ref{fig_wasserstein_triangle}, \ref{fig_wasserstein_triangle2} ,\ref{fig_wasserstein_triangle3} we show examples of interpolations of content images with the style images. We used in this experiment a coarse to fine approach, i.e starting from matching upper layers of VGG to lower layers.\\

 \section{Mixing Styles with Frechet Means and Optimal Transport Style Transfer}
\begin{figure}[ht] 
 \includegraphics{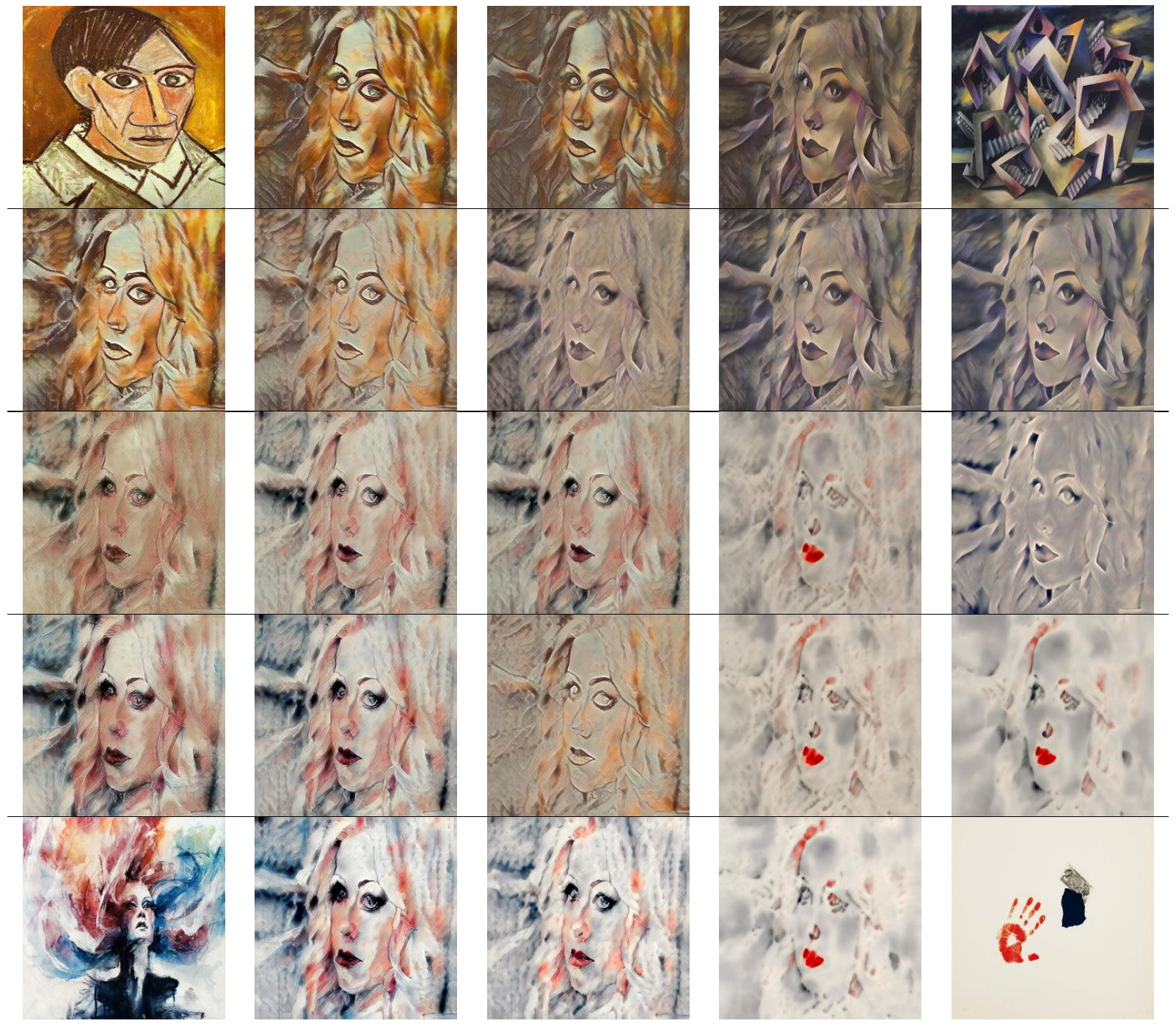}
 \caption{Coarse to Fine Generation: Karcher (Fisher Rao)  Barycenter Mixing of the styles and Wasserstein Transport of the content image to the novel style defined by the Fisher Rao barycenter. The stylized is generated following a coarse to fine scheme.  }
 \label{tab:tabFisherc2fapril}
 \end{figure}

\begin{figure}[ht] 
 \includegraphics{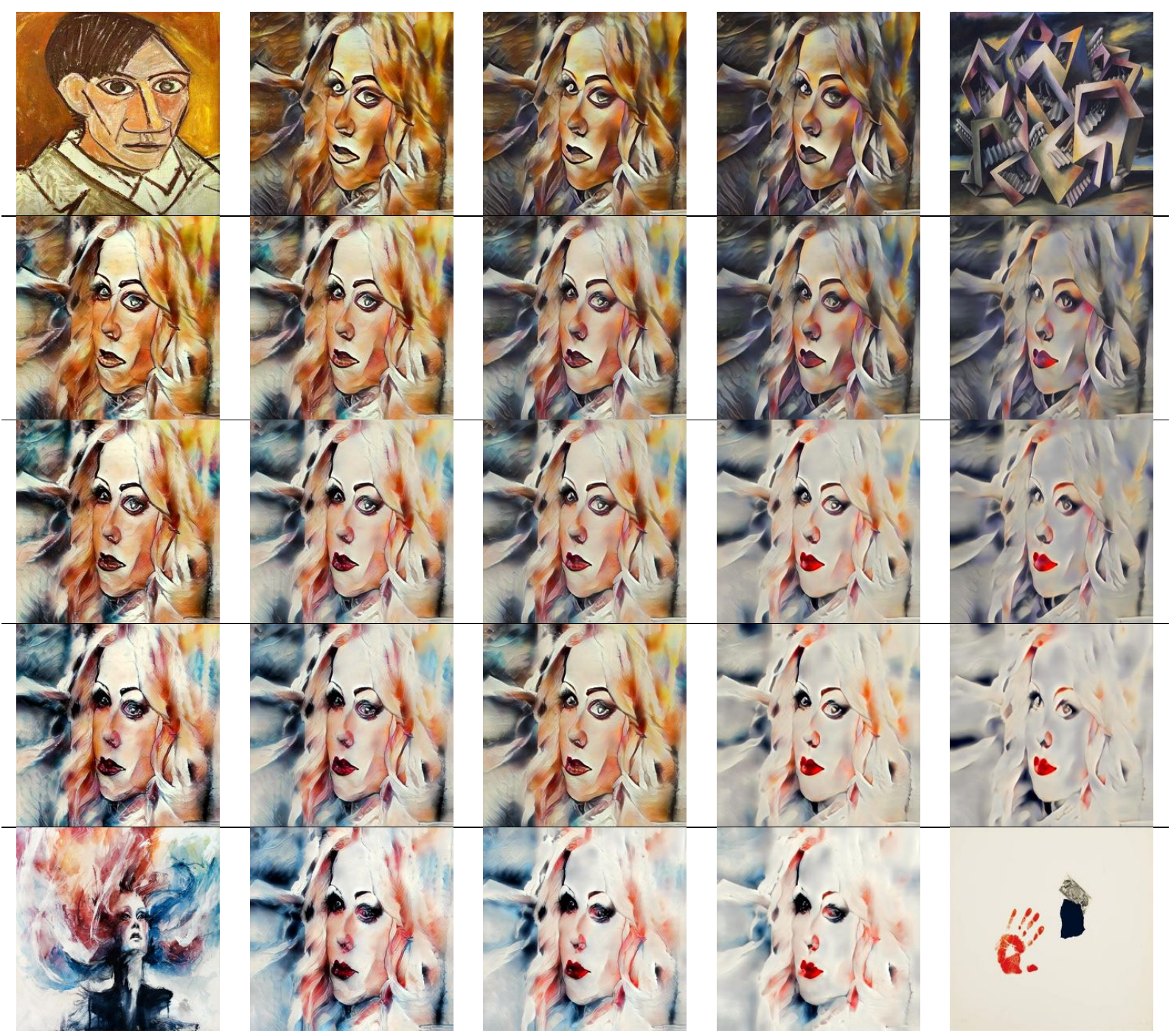}
 \caption{Coarse Fine Style Transfer:  (Arithmetic )Euclidean Barycenter Mixing  of Covariances and Wasserstein Transport. This is similar to a WCT type of mixing.  The stylized image is similar to a Wasserstein Barycenter Mixing, nevertheless a closer look shows subtle differences. This hints to the fact the coarse layers are almost diagonal.   } 
  \label{tab:tabArithmetic2fapril}
 \end{figure}

\begin{figure}[ht] 
 \includegraphics{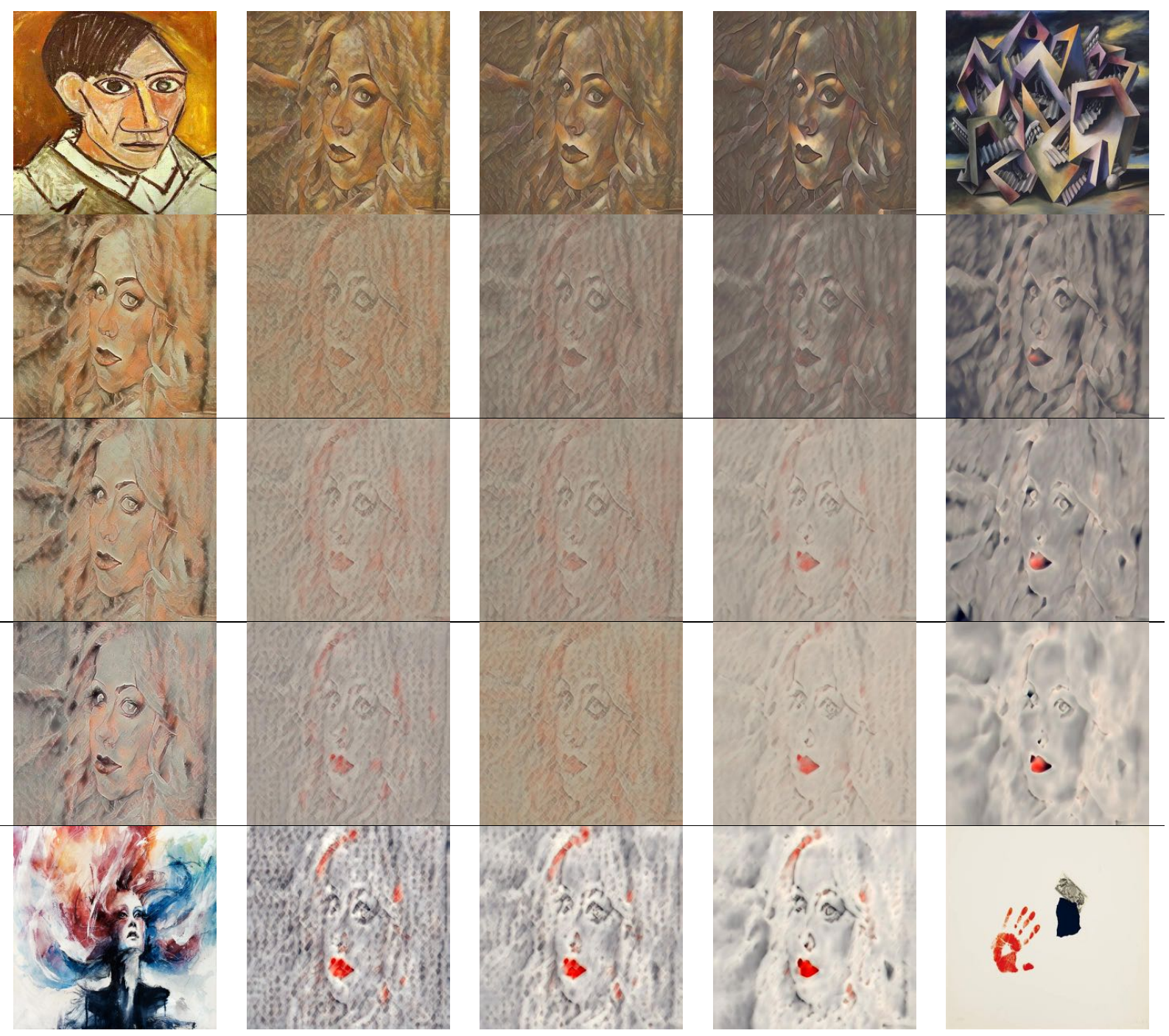}
 \caption{Coarse Fine Style Transfer:  Harmonic Barycenter Mixing  of Covariances and Wasserstein Transport. The Harmonic Mixing have saturation problems and does not produce good results. } 
\label{tab:tabHarmonic2fapril}
 \end{figure}

\begin{figure}[ht] 
 \includegraphics{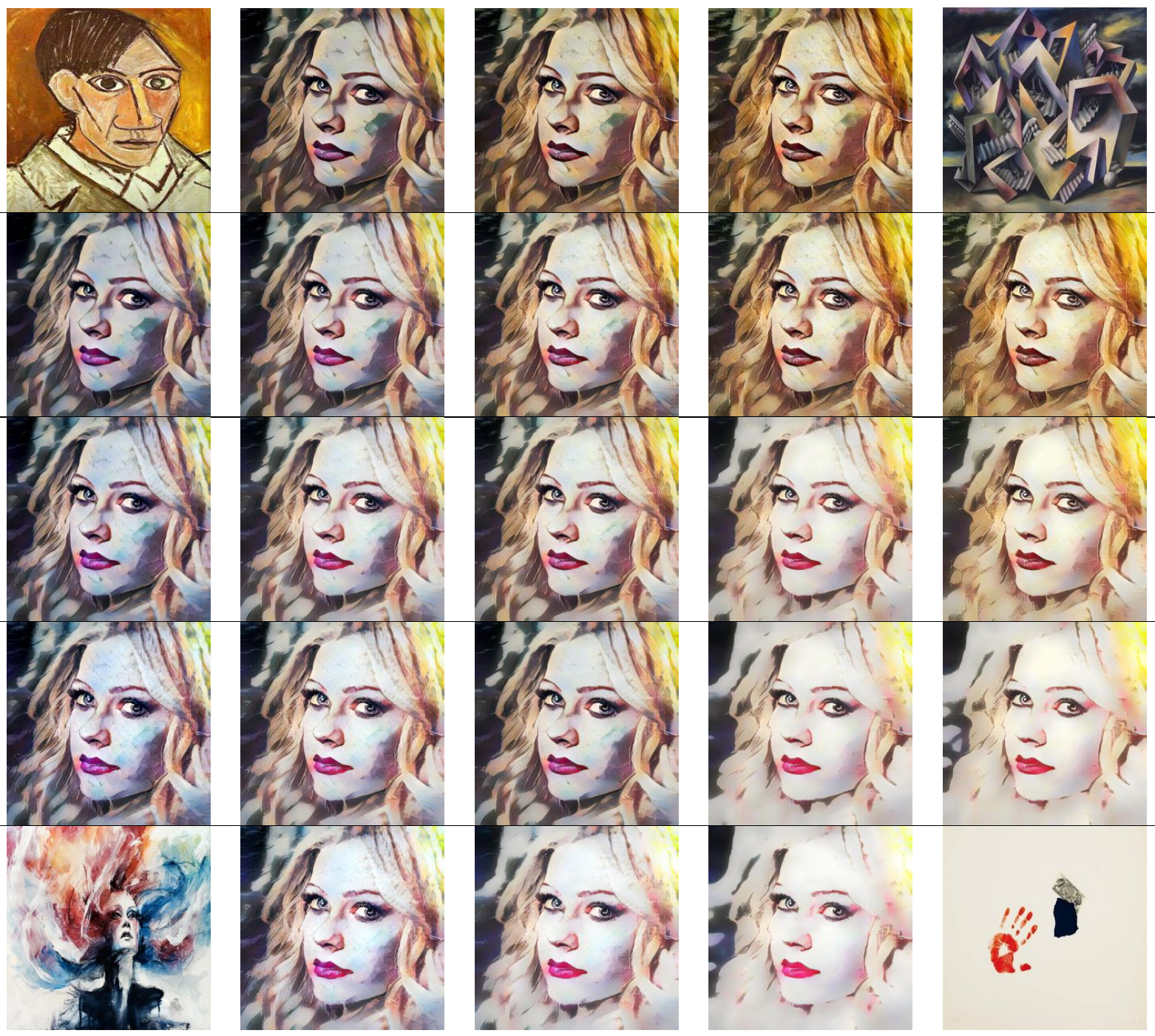}
 \caption{ Adaptive Instance Normalization Mixing Baseline \citep{AdaIn}. } \label{}
\label{tab:tabAdaINapril}
 \end{figure}

\begin{figure}[ht] 
 \includegraphics{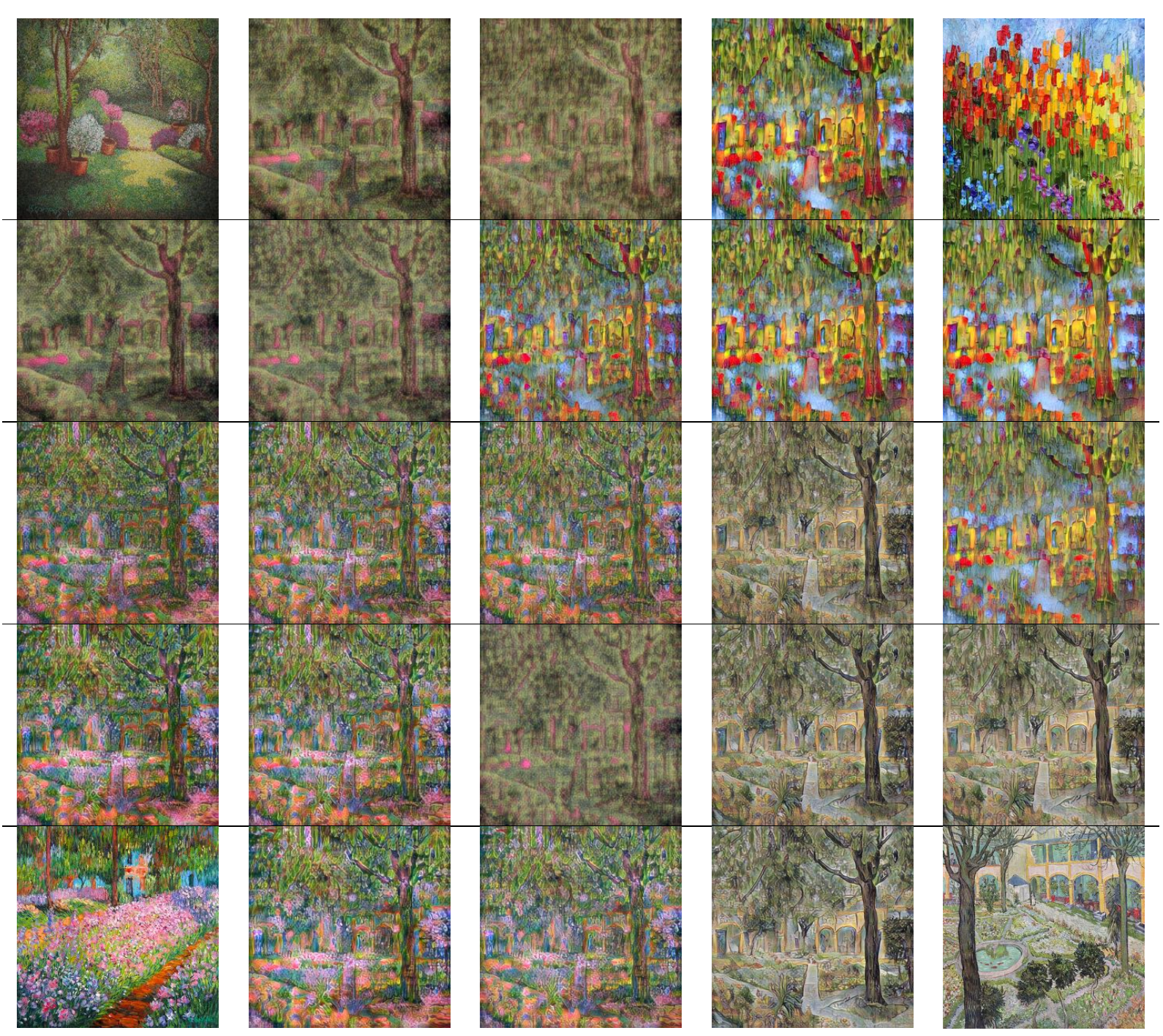}
  \caption{Coarse to fine style transfer: Fisher Rao  Mixing and Wasserstein Transport.  }
  \label{tab:vangogFisheRao}
 \end{figure}

\begin{figure}[ht] 
  \includegraphics{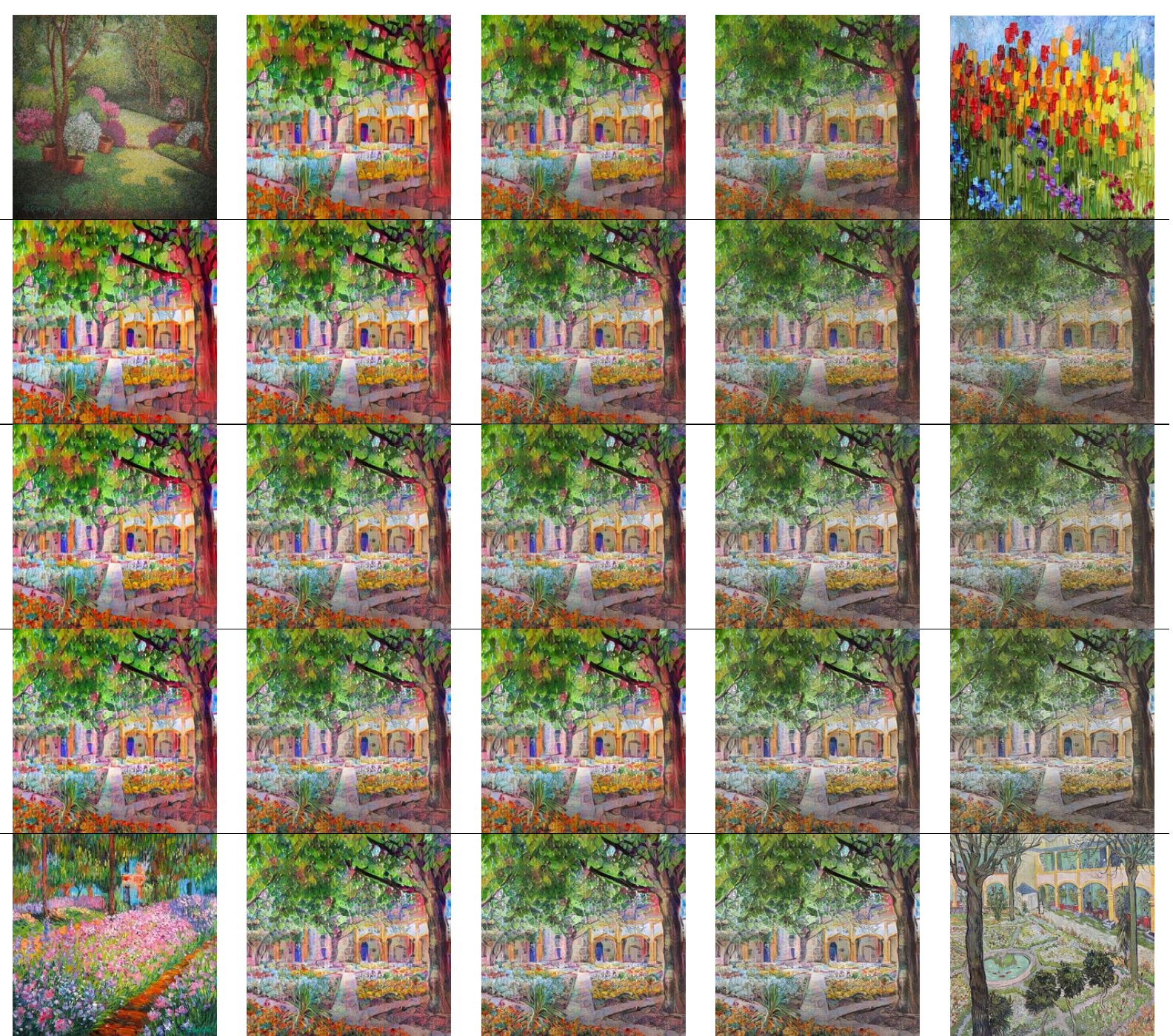}
 \caption{ Adaptive Instance Normalization Mixing Baseline \citep{AdaIn}. } \label{}
\label{tab:tabAdaINVANGOG}
 \end{figure}

\paragraph{Coarse to Fine.}
-We give results of different Mixing strategies and a content stylization in a coarse to fine procedure as follows: Wasserstein Mixing in Table \ref{tab:tabWassc2fapril};Fisher Rao Mixing in Table \ref{tab:tabFisherc2fapril} ;Arithmetic Mixing that would be close to WCT baseline \citep{WCT} in Table \ref{tab:tabArithmetic2fapril}; Harmonic Mixing in Table \ref{tab:tabHarmonic2fapril} ; AdaIN Mixing Table in \ref{tab:tabAdaINapril}. 
We also give another set of results on Wass Barycenter mixing in Table \ref{tab:VangogWass}, Fisher Rao in Table  \ref{tab:vangogFisheRao}  and AdaIn in \ref{tab:tabAdaINVANGOG}

\paragraph{Fine to Coarse.} We experiment baselining WCT mixing \citep{WCT} and Wasserstein Mixing in a Fine to coarse strategy (from lower layer to upper layers) results are given in  Table \ref{tab:tabWassc2faprilFLIP} and Table \ref{tab:tabWCTfaprilFLIP}.

 \begin{figure}[ht!]
 \centering
 \includegraphics[scale=0.15]{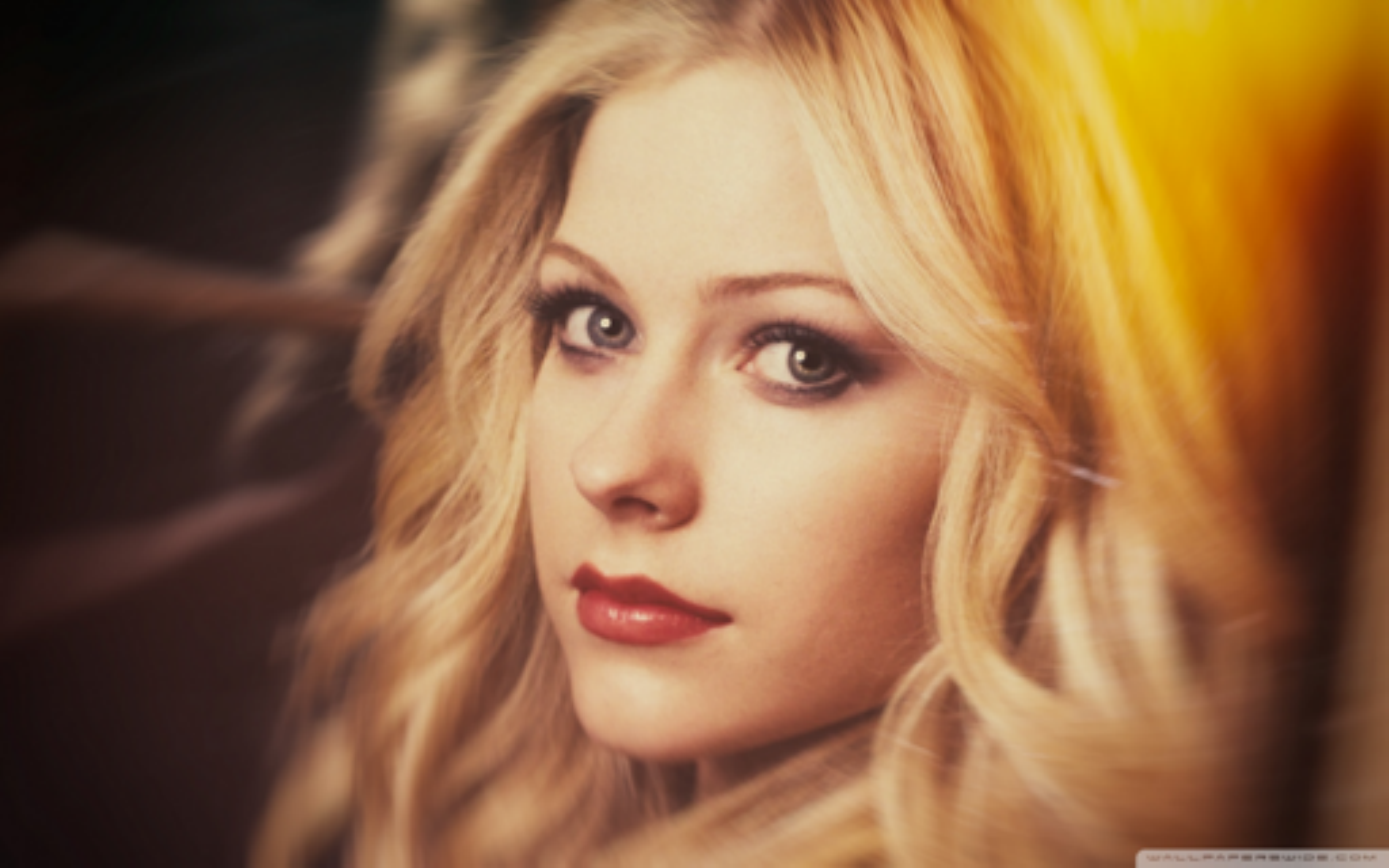}
 \caption{Content Image. We give results of different Mixing strategies and a content stylization in a coarse to fine procedure as follows: Wasserstein Mixing in Table \ref{tab:tabWassc2fapril};Fisher Rao Mixing in Table \ref{tab:tabFisherc2fapril} ;Arithmetic Mixing that would be close to WCT baseline \citep{WCT} in Table \ref{tab:tabArithmetic2fapril}; Harmonic Mixing in Table \ref{tab:tabHarmonic2fapril} ; AdaIN Mixing Table in \ref{tab:tabAdaINapril};  The style images are given on the four corners of each square.}
 \label{fig: april}
\end{figure}

\begin{figure}[ht] 
 \includegraphics{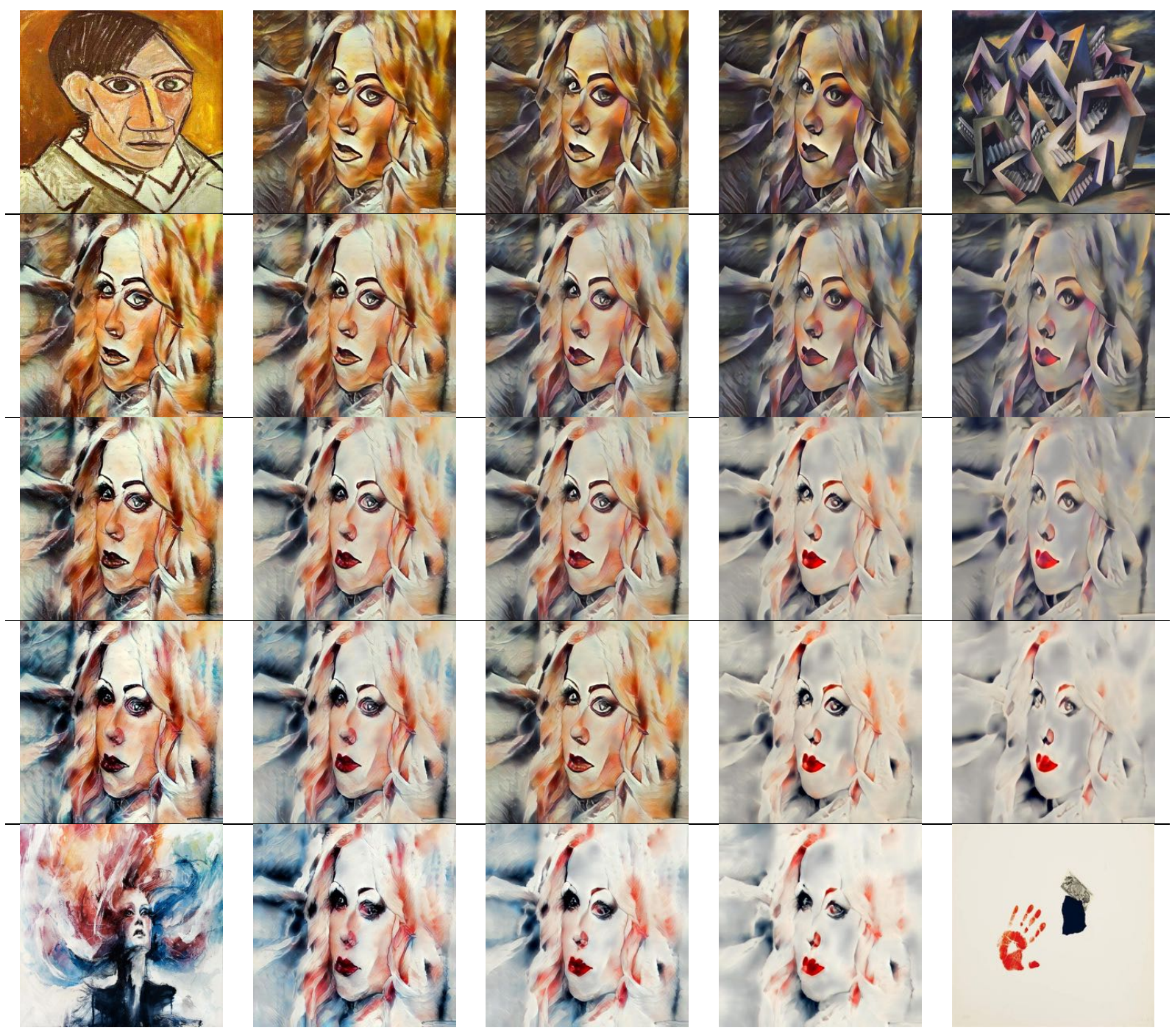}
 \caption{Coarse to Fine style Transfer: Content image is given in Figure \ref{fig: april}. Wasserstein Barycenter Mixing of the styles (the four images in the corners of the square) and Wasserstein Transport of the content image to the novel style defined by the Wasserstein Barycenter for various interpolations weights. The stylized image is generated by following a coarse to fine scheme.   } 
 \label{tab:tabWassc2fapril}

 \end{figure}

 \begin{figure}[ht] 
 \includegraphics{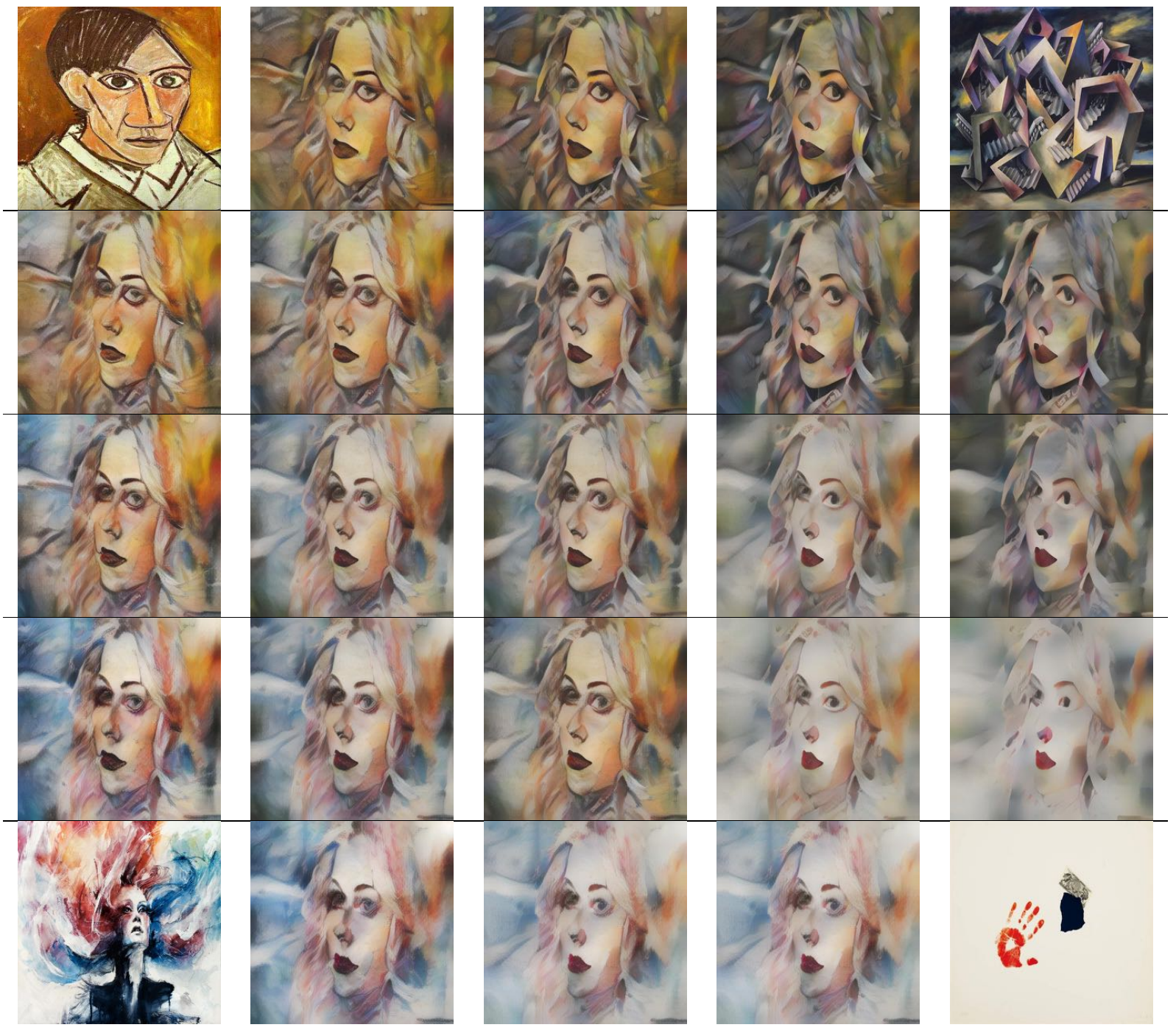}
 \caption{Fine to coarse style Transfer: Content image is given in Figure \ref{fig: april}. Wasserstein Barycenter Mixing of the styles (the four images in the corners of the square) and Wasserstein Transport of the content image to the novel style defined by the Wasserstein Barycenter for various interpolations weights. The stylized image is generated by following a coarse to fine scheme.   } 
 \label{tab:tabWassc2faprilFLIP}
 \end{figure}
 
  \begin{table}[ht] 
 \includegraphics{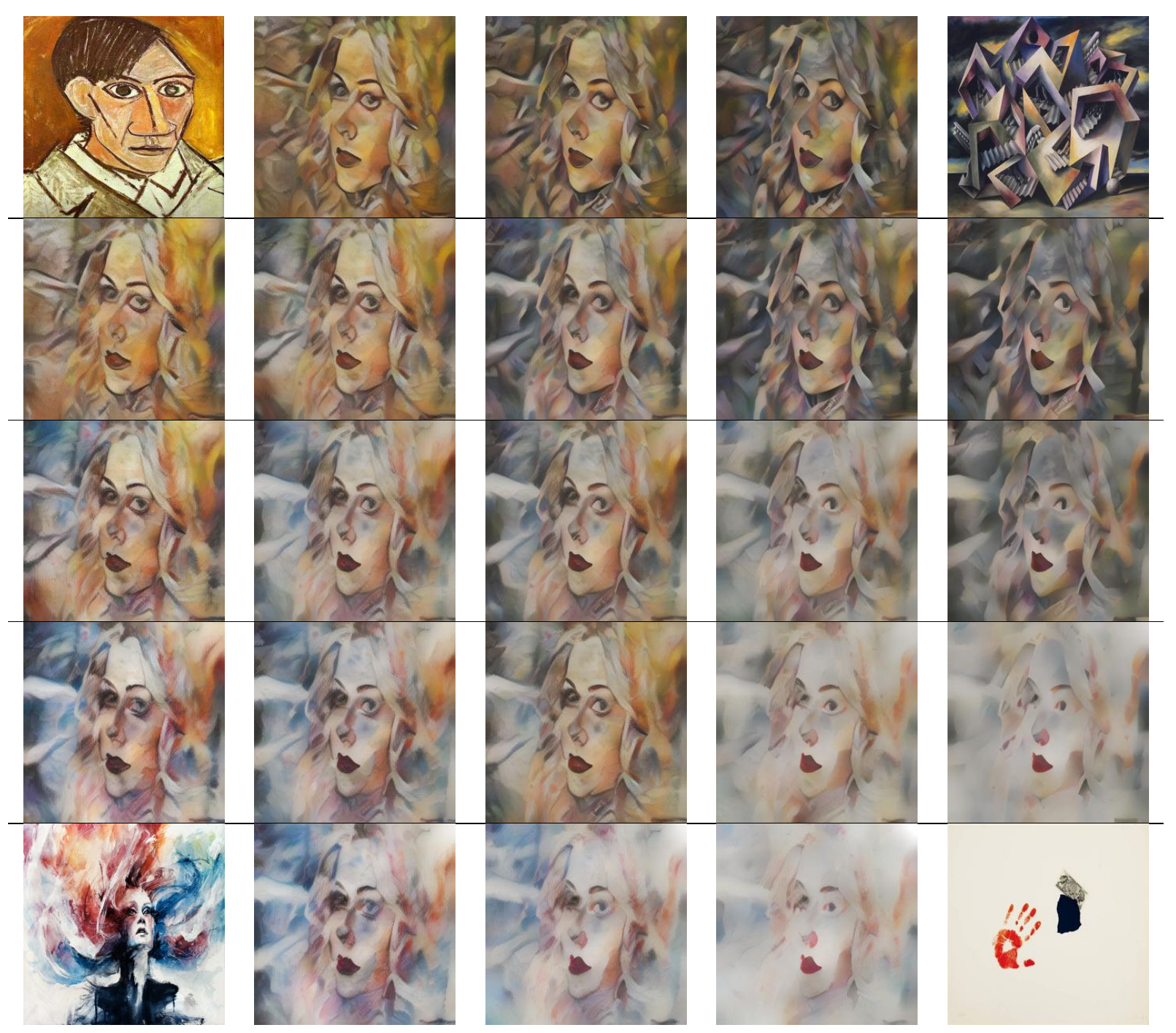}
 \caption{Fine to coarse style Transfer: Results of Arithmetic Mean Mixing (WCT Mixing \citep{WCT}) , we see here that the coloring has a lot of black shadow over the face unlike the Wasserstein barycenter mixing approach, in the previous Table. } 
 \label{tab:tabWCTfaprilFLIP}

 \end{table}

 \newpage
 \begin{figure}[ht!]
 \centering
 \includegraphics[scale=0.5]{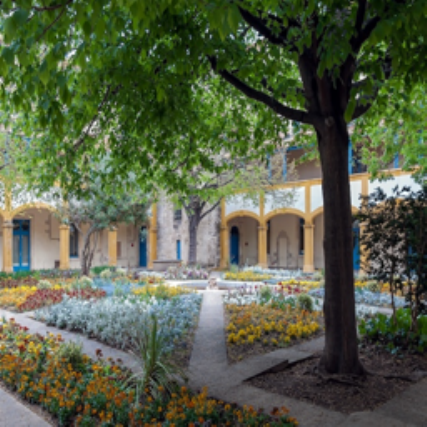}
 \caption{Content Image  ( a Photo of Hotel Dieu painted by van gogh  in the most right corner of the square).  Stylization in mixture of four styles incluing van gogh painting are in Tables \ref{tab:VangogWass} for Wasserstein mixing and Table \ref{tab:vangogFisheRao} for Fisher Rao Mixing. Table \ref{tab:tabAdaINVANGOG} is the AdaIN baseline.
}
\label{fig:HotelDieu}
 \end{figure}
 
 \begin{figure}[ht] 
 \includegraphics{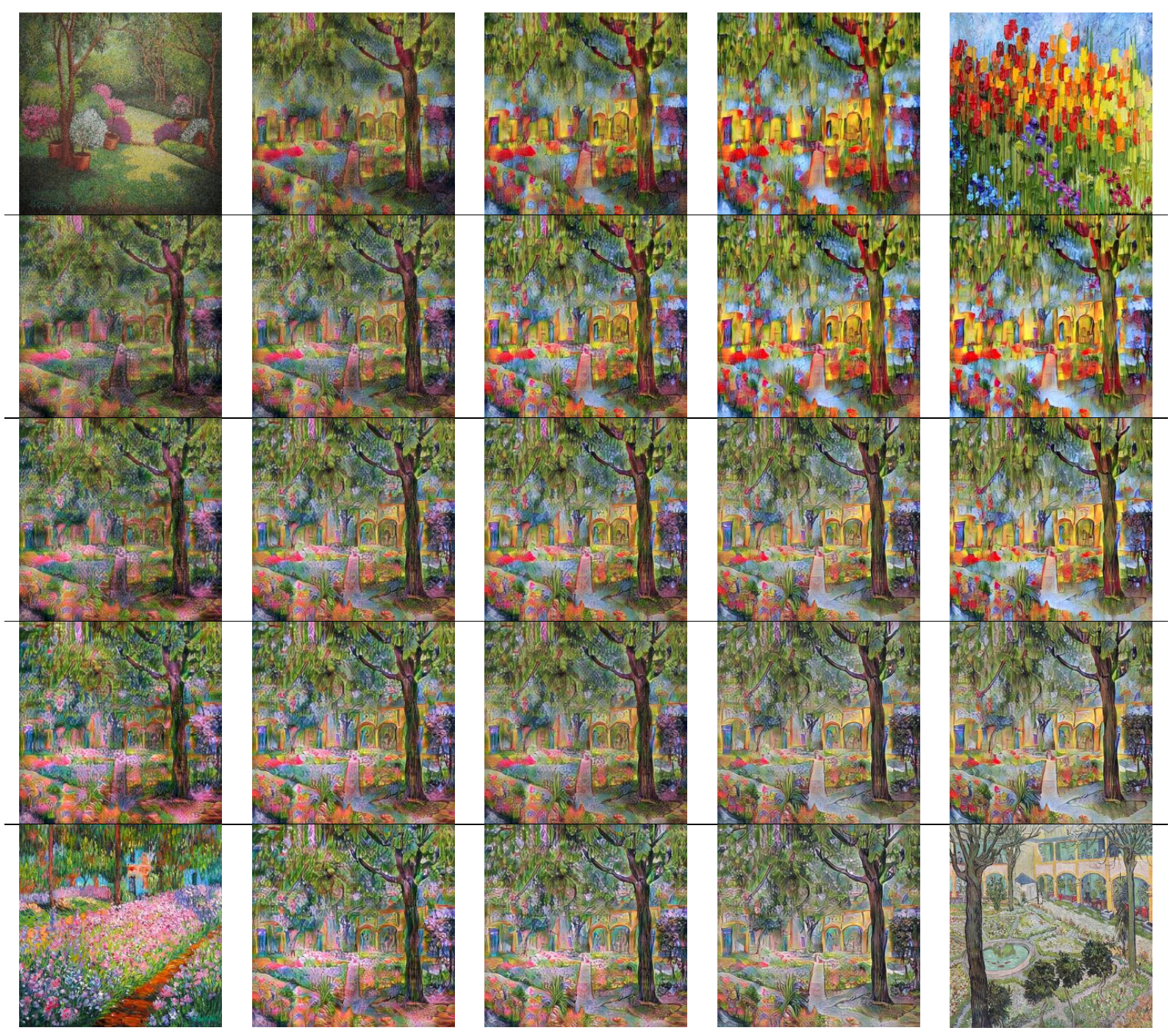}
 \caption{Coarse to Fine style transfer: Wasserstein Barycenter Mixing and Wasserstein Transport. The content image is given in Figure \ref{fig:HotelDieu}. The four styles are on the four corner of the square.  }
 \label{tab:VangogWass}
 \end{figure}
 
\end{document}